\documentclass[11pt, intlimits]{article}
\usepackage[letterpaper, margin=1.0in, footskip=30pt]{geometry}

\usepackage{amsmath}
\usepackage[T1]{fontenc}
\usepackage{mathrsfs}
\usepackage{graphicx}
\usepackage{pgfplots}
\usepackage{float}
\usepackage{tabu}
\usepackage{booktabs}
\usepackage{tikz}
\usepackage{tikz-cd}
\usepackage{url}
\graphicspath{{graphics/}}

\usepackage{caption}
\usepackage[backend=biber, sorting=none, style=numeric-comp]{biblatex}
\begin{document}

\thispagestyle{empty}

\begin{center}
{\LARGE Loss Terms and Operator Forms of Koopman Autoencoders}
\\
25 November 2024
\end{center}

\noindent
\begin{minipage}[t]{0.5\textwidth}
\begin{center}
Dustin Enyeart \\
Department of Mathematics \\
Purdue University \\
\url{denyear@purdue.edu} \\
\end{center}
\end{minipage}%
\hfill
\begin{minipage}[t]{0.5\textwidth}
\begin{center}
Guang Lin\footnotemark \\
Department of Mathematics \\
School of Mechanical Engineering \\
Purdue University \\
\url{guanglin@purdue.edu} 
\end{center}
\end{minipage}
\footnotetext{Corresponding author}

\begin{abstract}
\noindent Koopman autoencoders are a prevalent architecture in operator learning. 
But, the loss functions and the form of the operator vary significantly in the literature. 
This paper presents a fair and systemic study of these options. 
Furthermore, it introduces novel loss terms.  
\end{abstract}

\section{Introduction}
A \emph{neural operator}\index{neural operator} is a neural network that is intended to approximate an operator between function spaces \cite{kovachki2024operator, boulle2023mathematical, winovich2021neural}. 
An example of an output function for a neural operator is a solution to a differential equation. 
Examples of input functions for a neural operator are the initial conditions or the boundary conditions for the differential equation.
The study of neural operators is called \emph{operator learning}\index{operator learning}.

This paper is about Koopman autoencoders, which is a prevalent neural operator architecture to to learn the time evolution of differential equations \cite{koop1, mamakoukas2020learning, huang2020data, klus2020data}. 
They are especially popular\index{applications of Koopman architecture} for dynamic mode decomposition\index{dynamic mode decomposition} \cite{dymodedecomp1, dymodedecomp2, kutz2016multiresolution, takeishi2017learning} and control\index{control} \cite{budivsic2020koopman, li2019learning, kaiser2020data, han2020deep, bruder2021koopman, bruder2019modeling, arbabi2018data, abraham2017model}.
Other prevalent neural operator architectures are DeepONets \cite{DeepONet, lanthaler2022error, goswami2022physics, he2023novel, cho2024learning, lu2022comprehensive} and Fourier neural operators \cite{li2020fourier, qin2024toward, li2023fourier, brandstetter2022lie, kovachki2021universal, tran2021factorized, tancik2020fourier, lu2022comprehensive}. 

In the literature, the loss functions and operator forms for Koopman autoencoders vary significantly, and there is not a systematic comparison of them. 
This chapter presents a fair and comprehensive comparison between loss functions and operator forms for Koopman autoencoders.
Furthermore, it introduces novel loss terms that are original to this dissertation. 

First, Koopman autoencoders are introduced in Section~\ref{sec-architecture}.
And, the loss functions and operator forms are introduction in Section~\ref{sec-loss}.
The differential equations used in this paper are introduced in Section~\ref{sec-diffeq}.
Then, the loss functions and operator forms are introduction in Section~\ref{sec-loss}.
Then, large grid-search experiments are presented in Section~\ref{sec-large}.
These experiments are over many combinations for only a small number of epochs. 
The purpose of these experiments is to identify promising loss functions through robust trends.  
The purpose of these experiments is to identify robust trends to identify the most promising loss functions.
Then, in Section~\ref{sec-operator}, the operator forms and loss functions for the operator forms are compared in more detail.
Finally, this paper ends with an overall conclusion in Section~\ref{sec-conclusion}.

\section{Koopman Autoencoders}\label{sec-architecture}
The Koopman formulation of classical mechanics\index{Koopman formulation of mechanics} is an alternative framework of classical mechanics \cite{koopman1931hamiltonian, brunton2021modern, bruce2019koopman}. 
It was inspired by the Hamiltonian formulation of quantum mechanics \cite{morrison1990understanding, ohanian1989principles, griffiths2018introduction}. 
In this theory, the physical state\index{physical space} at a given time is represented as a state in an infinite-dimensional latent space\index{latent space}, and an infinite-dimensional operator governs the time evolution in this latent space.
Intuitively, the Koopman formulation can be thought of as removing nonlinearity by infinitely increasing the dimension. 

The Koopman formulation of classical mechanics can be discretized\index{discretized Koopman formulation} to provide a numerical scheme by approximating the time-evolution operator by a finite-dimensional matrix $K$.
In this scheme, a physical state $s_0$ can be evolved into a later physical state $s_n$ by encoding it into the latent space, applying the matrix $K$ repetitively and then decoding it back to the physical space, that is, the equation 
\[
    s_n = R \circ K^{n} \circ E(s_0)
\]
approximately holds, where $E$ is a discretized encoder and $R$ is a discretized decoder.
In this numerical scheme, the dimension of this latent space is called the \emph{encoding dimension}\index{encoding dimension}.

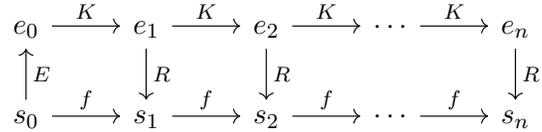
\begin{figure}[H]
\[
\begin{tikzcd}
    e_0 \arrow{r}{K}
    & e_1 \arrow{d}{R} \arrow{r}{K}
    & e_2 \arrow{d}{R} \arrow{r}{K}
    & \cdots \arrow{r}{K}
    & e_n \arrow{d}{R} \\
    s_0 \arrow[swap]{u}{E} \arrow{r}{f}
    & s_1 \arrow{r}{f}
    & s_2 \arrow{r}{f}
    & \cdots \arrow{r}{f}
    & s_n \\
\end{tikzcd}
\]
\caption{Discretization of the Koopman formulation into a numerical scheme: The physical states at successive time points are denoted by $s_0$, $s_1$, $\dots$, $s_{n-1}$ and $s_n$. The encoded states at successive time points are denoted by $e_0$, $e_1$, $\dots$, $e_{n-1}$ and $e_n$. The function $f$ is the true time evolution of the physical state by the time step. The discretized Koopman operator, encoder and decoder are denoted by $K$, $E$ and $R$, respectively.}
\end{figure}    

\emph{Koopman autoencoders}\index{Koopman autoencoder}\index{Koopman architecture}\index{Koopman neural network} are based on the discretized Koopman formulation where the operator\index{Koopman operator}, encoder\index{Koopman encoder} and decoder\index{Koopman decoder} are neural networks.
The encoder and decoder normally consists of fully connected layers or, sometimes for partial differential equations, convolution layers. 
And, the operator is a fully connected layer without a bias, and an activation function\index{activation function} is not used between successive applications of it.  
Because the operator is applied repeatedly, it can be useful to use gradient clipping.

\section{Loss Functions and Operator Forms}\label{sec-loss} 
Intuitively, during training, the loss function\index{loss function} for the Koopman architecture should regulate three parts. 
First, the loss function needs to regulate the accuracy of the model, that is, the differences between the predictions and the targets. 
Second, the loss function should regulate the correspondence between the encoder and decoder because the decoding of an encoding should be the original input. 
Third, the loss function should regulate the Koopman operator so that it is close to being unitary because the Koopman operator in classical mechanics is a unitary operator.
These three types of loss terms are called the \emph{accuracy loss}\index{accuracy loss} terms, \emph{encoding loss}\index{encoding loss} terms and \emph{operator loss}\index{operator loss} terms, respectively.

\subsection{Accuracy Loss}

The accuracy loss terms measure how well the model predicts the physical states. The accuracy loss terms in this paper are the full accuracy loss, max loss and discounted accuracy loss.

The \emph{full accuracy loss}\index{full accuracy loss} is the mean squared error between the predictions and the targets, that is, it is the number
\[
    \frac{1}{n} \cdot \sum_i \vert \vert R \circ K^i \circ E(v_0) - v_i \vert \vert^2
    ,
\]
where $K$ is the Koopman operator, $E$ is the encoder, $R$ is the decoder, $n$ is the number of time steps and $v_i$ is the $i$-th physical state.
This is the most fundamental loss function, and it is what is used for testing. 

The \emph{max loss}\index{max loss} is the maximum mean squared error between the predictions and the targets, that is, it is the number
\[
    \max_i \vert \vert R \circ K^i \circ E(v_0) - v_i \vert \vert^2
    , 
\]
where $K$ is the Koopman operator, $E$ is the encoder, $R$ is the decoder and $v_i$ is the $i$-th physical state \cite{lusch2018deep, shi2022deep}.
It can be used by itself or with the full accuracy loss to weigh the worse prediction more. 

The \emph{discounted accuracy loss}\index{discounted accuracy loss} is a weighted mean squared error between the predictions and the targets such that earlier time steps are weighed more. 
The idea is to prevent the model from getting stuck in at a local minimum such as a steady state. 
This is an idea from reinforcement learning \cite{winder2020reinforcement}.  
In this paper, this is done by weighing the $i$-th time step by $\lambda^i$, where $\lambda$ is a positive real number that is less than $1$. The number $\lambda$ is called the \emph{discount factor}\index{discount factor}. 
Thus, the discounted accuracy loss is the number
\[
    \frac{1}{n} \cdot \sum_i \lambda^i \vert \vert R \circ K^i \circ E(v_0) - v_i \vert \vert^2
    ,
\]
where $K$ is the Koopman operator, $E$ is the encoder, $R$ is the decoder, $n$ is the number of time steps and $v_i$ is the $i$-th physical state.
This is a generalization of the full accuracy loss in the sense that the full accuracy loss is the discounted accuracy loss with the discount factor $1$.
The authors are unaware of any papers that use such a loss, that is, it is novel to this paper for the Koopman architecture.

\subsection{Encoding Loss}

The encoding loss terms restrain the encoder and decoder. 
Intuitively, the decoding of an encoding should be the original input.
The encoding loss terms in this paper are the reconstruction loss, the consistency loss and the metric loss. 

The \emph{reconstruction loss}\index{reconstruction loss} is the mean-square error between the vectors in physical space and their images after applying both the encoder and the decoder, that is, it is the number
\[
    \frac{1}{n} \cdot \sum_{i} \vert \vert R \circ E(v_i) - v_i \vert \vert^2
    ,
\]
where $E$ is the encoder, $R$ is the encoder, $n$ is the number of time steps and each $v_i$ is a vector in the physical space \cite{lusch2018deep, li2021deep}. 
Intuitively, it encourages the composition of the decoder and the encoder to be the identity
In the numerical experiments in this paper, the physical states were used, but random vectors could also be used.

The \emph{consistency loss}\index{reconstruction loss} is the mean squared error between the predicted encoded state at a time and the encoding of the target physical state, that is, it is the number
\[
    \frac{1}{n} \cdot \sum_{i} \vert \vert K^i \circ E(v_0) - E v_i \vert \vert^2, 
\]
where $K$ is the Koopman operator, $E$ is the encoder, $n$ is the number of time steps and each $v_i$ is a vector in the physical space \cite{rostamijavanani2023study}.
Intuitively, this together with an accuracy term encourages the compositions of the decoder and the encoder to be the identity.

The \emph{metric loss}\index{metric loss} is the mean squared error between the squared distances between encoded states and physical states, that is, it is the number
\[
    \frac{1}{n} \cdot \sum_{i} 
    \big\vert \big\vert
    \vert \vert E(v_i) - E(v'_i) \vert \vert^2
    - \vert \vert v_i - v'_i \vert \vert^2
    \big\vert \big\vert^2 
    ,
\]
where $E$ is the encoder, each $v_i$ and $v'_i$ are vectors in the physical space and $n$ is the number of vector pairs \cite{li2019learning}.
In the Koopman formulation of classical physics, the distance between physical states is equal to the distance between their corresponding encoded states. 
Thus, intuitively, this encourages the structure of the physical states and the geometry of the encoded states to be related.

\subsection{Operator Form and Operator Loss}

Because the operator in the koopman formulation is unitary, it is intuitively reasonable to use a loss function that encourages the operator to be unitary.
Such a term loss term would measure the failure of the operator to be unitary. 
The operator loss terms in this paper are the isometry loss, norm loss, unitary loss and determinant loss.
Furthermore, because the latent space for Koopman autoencoders is learned, it can be restricted so that its matrix has a certain form in the basis of the latent space. 
The different operator forms in this paper are the dense form, the tridiagonal form and the Jordan form. 

If there is not a restriction on the operator, then all entries of its matrix are learnable parameters. 
This is the \emph{dense form}\index{dense form} of the operator.

Because, a unitary operator is tridiagonal in some basis, the matrix for the operator can be restricted to be tridiagonal, that is, it is of the form
\[
\begin{pmatrix}
    a_1 & b_1 \\
    c_1 & a_2 & b_2 \\
    & c_2 & \ddots & \ddots \\
    & & \ddots & \ddots & b_{n-1} \\
    & & & c_{n-1} & a_n
\end{pmatrix}
\]
where, for each index $a_i$, $b_i$ and $c_i$ are real numbers.
This is called the \emph{tridiagonal form}\index{tridiagonal form} of the operator. 
To replicate this structure, a mask\index{Koopman mask} can be place on the operator in the Koopman architecture to make it tridiagonal.

In some basis, a real matrix can be written as a block diagonal matrix where each block is $1 \times 1$ or $2 \times 2$ of the form
\[
\begin{pmatrix}
    a & b \\
    -b & a
\end{pmatrix},
\]
where $a$ and $b$ are real numbers \cite{johnson1985matrix}. 
This is the \emph{Jordan form}\index{Jordan form} of the operator.
In this paper, this is implemented using only $2 \times 2$ blocks, that is, the matrix is written in the form 
\[
\begin{pmatrix}
    a_1 & b_1 & 0 & 0 \\
    -b_1 & a_1 &  0 & 0 
    \\
    0 & 0 & a_2 & b_2 \\
    0 & 0 & -b_2 & a_2 \\
    & & & & \ddots & \ddots & 0 & 0\\
    & & & & \ddots & \ddots & 0 & 0 \\
    & & & & 0 & 0 & a_n & b_n \\
    \\
    & & & & 0 & 0 & -b_n & a_n \\
\end{pmatrix},
\]
where, for each index $i$, $a_i$ and $b_i$ are real numbers \cite{lusch2018deep}. 
Intuitively, because this matrix already represents a learned discretized version of a continuous operator, using only $2 \times 2$ blocks will be sufficiently expressive because, at worse, it will expressive redundant information. 
Similarly to the tridiagonal restriction, this reduces the number of learnable parameters. 
Furthermore, this structure allows for the eigenvalues to be easily computed because the eigenvalues are of the form $a_i + b_i \cdot \mathrm{i}$. This is useful in dynamic mode decomposition.

The \emph{norm loss}\index{norm} is the squared distance between the square of the $\mathrm{L}^2$-norm of the matrix and $1$, that is, it is the number
\[
    \big\vert \big\vert 
    \vert \vert K \vert \vert^2 - 1
    \big\vert \big\vert^2 
    , 
\]
where $K$ is the Koopman operator \cite{khosravi2023representer}. 
The intuition of this loss term is that the norm of a unitary operator is $1$ \cite{trefethen2022numerical}.

The \emph{isometry loss}\index{isometry loss} is the mean squared error between the squares of the norms of vectors and the squares of the norms of their encodings, that is, it is the number
\[
    \frac{1}{n} \cdot \sum_{i} \big\vert \vert \vert K(v_i) \vert \vert^2 - \vert \vert v_i \vert \vert^2 \big\vert^2
    ,
\]
where $K$ is the Koopman operator, $n$ is the number of vectors considered and each $v_i$ is a vector in the physical space. 
The intuition of loss term is that a unitary operator preserves the norm \cite{trefethen2022numerical}. This can be done with randomly generated data or with data used for training. In the experiments in this paper, randomly generated data was used. 

The \emph{unitary loss}\index{unitary loss} is the mean squared error between $K \circ K^{\mathrm{T}}$
and the identity matrix, that is, it is the number 
\[
    \vert \vert K \circ K^{\mathrm{T}} - I \vert \vert^2
    ,
\]
where $K$ is the Koopman operator. 
The intuition of loss term is that the adjoint of a unitary operator is its inverse \cite{trefethen2022numerical}. 
The authors are unaware of any papers that use this loss term, that is, this term is novel to this paper.

The \emph{determinant loss}\index{determinant loss} is the squared distance from the determinant of the Koopman operator to $1$, that is, it is the number
\[
    \vert \vert \det(K) - 1 \vert \vert^2
    ,
\]
where $K$ is the Koopman operator. 
The intuition of this loss term is that the determinant of a unitary operator is $1$ or $-1$ \cite{trefethen2022numerical}.
Computing the determinant of a matrix whose size is that of the Koopman operator is usually too computationally expensive for it to be practical. 
But, if a matrix is tridiagonal, then the determinant is computationally cheap because the determinant of a tridiagonal matrix
\[
\begin{pmatrix}
    a_1 & b_1 \\
    c_1 & a_2 & b_2 \\
    & c_2 & \ddots & \ddots \\
    & & \ddots & \ddots & b_{n-1} \\
    & & & c_{n-1} & a_n
\end{pmatrix}
\]
can be computed by the formula
\[
    \mathrm{det}(A_{m}) 
    = a_m\mathrm{det}(A_{m-1}) - b_{m-1}c_{m-1}\mathrm{det}(A_{m-2})
    , 
\]
where $A_i$ denotes the submatrix $A[1 \! : \! i,\ 1 \! : \! i]$ of $A$. 
Furthermore, in the Jordan form, the identity blocks on the upper diagonal do not affect its determinant, and the determinant can be computed as if the matrix is tridiagonal.
The authors are unaware of any papers that use this loss term, that is, this term is novel to this paper.

For time-reversible differential equation, two koopman operators can be trained where one is for stepping forward through time and the other is for stepping backward through time.
Then, a loss term that enforces the inverse operator to be the inverse of the Koopman operator can be implemented.
This is called the \emph{reversibility loss}\index{reversibility loss} \cite{azencot2020forecasting}.
This loss term is not experimented with in this paper.

\subsection{Auxiliary Loss}

The \emph{absolute max loss}\index{absolute max loss} is the maximum mean squared error between the predictions and the targets across all sample points $x_j$, that is, it is the number
\[
    \max_i \max_j
    \vert \vert (R \circ K^i \circ E(v_0))_j - v_{i,j} \vert \vert^2
    , 
\]
where $K$ is the Koopman operator, $E$ is the encoder, $R$ is the decoder, $v_{i, j}$ is the $i$-th physical state at the sample point $x_j$.
While the max loss is the maximum mean squared error across the time slices, the absolute max loss is the maximum mean squared error across the time slices and the sample points. 
The max loss and the absolute max loss are the same for ordinary differential equations, but they are different for partial differential equations. 
By itself, in practice, the absolute max loss is not enough to be an accuracy loss, but it can be useful together with another accuracy term.
Intuitively, it penalizes the most extreme errors. 
The authors are unaware of any papers that use this loss term, that is, it is novel to this paper. 

Physics-informed loss \index{physics-informed loss} terms are terms that encourage the model to obey the laws of physics \cite{liu2024physics, rice2020analyzing}.
An example is enforcing the model to conserve energy in systems where energy is conserved, such as for simple harmonic motion, the pendulum or wave equation. 
Another example of a physics-informed loss term is enforcing Gauss's laws in electromagnetism.
The energy-conservation loss is experimented with in the numerical experiments for the pendulum.

\section{Differential Equations}\label{sec-diffeq}
This section introduces the differential equations that are used for the numerical experiments in this paper.
The ordinary differential equations are the equation for simple harmonic motion, the equation for the pendulum, the Lorenz system and a fluid attractor equation. 
The partial differential equations are the wave equation, the heat equation, Burger's equation and the Korteweg-de-Vries equation.

\subsection{Simple Harmonic Motion}

The equation for \emph{simple harmonic motion}\index{simple harmonic motion} is the differential equation
\[
    \frac{\mathrm{d}^2 x}{\mathrm{d}^2 t}
    = - x
    .
\]
This is a time-dependent second-order linear ordinary differential equation whose dimension is $1$. 
This equation is used to model a mass on a spring, where the variable $x$ is the displacement of the mass from its equilibrium position.

In the numerical experiments in this paper, the models attempt to learn the solution to this differential equation as a function of the initial condition. 
To get data, initial positions are generated randomly.
Then, this differential equation is numerically solved for these initial conditions using the Runge-Kutta Method \cite{leveque2007finite}.

\subsection{Pendulum}

The equation for the \emph{pendulum}\index{pendulum} is the differential equation
\[
    \frac{\mathrm{d}^2 \theta}{\mathrm{d}^2 t}
    = - \sin (\theta)
    . 
\]
This is a time-dependent second-order nonlinear ordinary differential equation whose dimension is $1$. 
This equation models the motion of a pendulum in a constant gravitational field, where the variable $\theta$ is the angle of the pendulum from vertical.

In the numerical experiments in this paper, the models attempt to learn the solution to this differential equation as a function of the initial condition.
To get data, initial positions are generated randomly.
Then, this differential equation is numerically solved for these initial conditions using the Runge-Kutta Method \cite{leveque2007finite}.

\subsection{Lorenz System}

The \emph{Lorenz system}\index{Lorenz system} is the differential equation
\[
\begin{cases}
    \frac{\mathrm{dx}}{\mathrm{d}t} = y - x  \\
    \frac{\mathrm{dy}}{\mathrm{d}t} = x - x \cdot z - y  \\
    \frac{\mathrm{dz}}{\mathrm{d}t} = x \cdot y - z. \\ 
\end{cases}
\]
This is a time-dependent first-order nonlinear ordinary differential equation whose dimension is $3$. 
Historically, this equation was used in weather modeling.
Now, it provides a benchmark example of a chaotic system\index{chaotic system}.

In the numerical experiments in this paper, the models attempt to learn the solution to this differential equation as a function of the initial condition.
To get data, initial positions are generated randomly.
Then, this differential equation is numerically solved for for these initial conditions using the Runge-Kutta Method \cite{leveque2007finite}.

\subsection{Fluid Attractor Equation}

The differential equation 
\[
\begin{cases}
    \frac{\mathrm{dx}}{\mathrm{d}t} = x - y + x \cdot z  \\
    \frac{\mathrm{dy}}{\mathrm{d}t} = x + y + y \cdot z \\
    \frac{\mathrm{dz}}{\mathrm{d}t} = x^2 + y^2 + z \\ 
\end{cases}
\]
is used to model fluid flow around a cylinder\index{fluid attractor equation} \cite{noack2003hierarchy}. 
It is a time-dependent first-order nonlinear ordinary differential equation whose dimension is $3$.

In the numerical experiments in this paper, the models attempt to learn the solution to this differential equation as a function of the initial condition.
To get data, initial positions are generated randomly.
Then, this differential equation is numerically solved for for these initial conditions using the Runge-Kutta Method \cite{leveque2007finite}.

\subsection{Heat Equation}

The \emph{heat equation}\index{heat equation} is the differential equation is 
\[
    \frac{\partial u}{\partial t} = \frac{\partial^2u}{\partial ^2x}
    .
\]
This equation is a time-dependent first-order linear partial differential equation whose range dimensions are both $1$. 
It is used to model heat flow. 

In the numerical experiments in this paper, the models attempt to learn the solution to these differential equations as a function of the initial condition.
Furthermore, Dirichlet boundary conditions are used for these equations such that the value of the unknown function is $0$ on the boundary.
To get data, initial conditions are made by generating random symbolic expressions that satisfy the boundary conditions, and these expressions are then numerically sampled. 
Then, the partial differential equation is numerically solved for these initial conditions \cite{langtangen2017finite}.

\subsection{Wave Equation}

The \emph{wave equation}\index{wave equation} is the differential equation is 
\[
    \frac{\partial^2 u}{\partial^2 t} = \frac{\partial^2 u}{\partial^2 x}
    .
\] 
This equation is a time-dependent second-order linear partial differential equation whose range dimensions are both $1$. 
It is used to model waves on strings. 

In the numerical experiments in this paper, the models attempt to learn the solution to these differential equations as a function of the initial condition.
Furthermore, Dirichlet boundary conditions are used for these equations such that the value of the unknown function is $0$ on the boundary.
To get data, initial conditions are made by generating random symbolic expressions that satisfy the boundary conditions, and these expressions are then numerically sampled. 
Then, the partial differential equation is numerically solved for these initial conditions \cite{langtangen2017finite}.

\subsection{Burger's Equation}

\emph{Burger's equation}\index{Burger's equation} is the differential equation
\[
    \frac{\partial u}{\partial t}
    = \frac{\partial^2 u}{\partial^2 x} - u\frac{\partial u}{\partial x} 
    .
\]
This is a time-dependent first-order nonlinear partial differential equation whose domain dimension and range dimension are both $1$. 
It is used to model some fluids. 

In the numerical experiments in this paper, the models attempt to learn the solution to this differential equation as a function of the initial condition.
Furthermore, Dirichlet boundary conditions are used for this equation such that the value of the unknown function is $0$ on the boundary.
To get data, initial conditions are made by generating random symbolic expressions that satisfy the boundary conditions, and these expressions are then numerically sampled. 
Then, the partial differential equation is numerically solved for these initial conditions \cite{leveque1992numerical}.

\subsection{Korteweg-de-Vries Equation}

The \emph{Korteweg-de-Vries equation}\index{Korteweg-de-Vries equation}, which is abbreviated as the \emph{KdV equation}\index{KdV equation}, is the differential equation
\[
    \frac{\partial u}{\partial t}
    = 6u\frac{\partial u}{\partial x}
    - \frac{\partial^3u}{\partial ^3x}
    .
\]
This is a time-dependent first-order nonlinear partial differential equation whose domain dimension and range dimension are both $1$. 
It is used to model some fluids. 

In the numerical experiments in this paper, the models attempt to learn the solution to this differential equation as a function of the initial condition.
Furthermore, periodic boundary conditions are used for this equation, that is, the value of the unknown function is the same on each endpoint of the spatial domain. 
To get data, initial conditions are made by generating random symbolic expressions that satisfy the boundary conditions, and these expressions are then numerically sampled. 
Then, the partial differential equation is numerically solved for these initial conditions \cite{zabusky1965interaction}.

\section{Large Grid Searches}\label{sec-large}
The experiments in this section are grid searches that cover a large number of combinations for a only a small number of epochs. 
The objective of these numerical experiments is to find robust trends in the choice of loss terms and options. 
This is done by analyzing mean effects, the best combinations and direct comparisons. 

For each numerical experiment, minimal preliminary testing was performed to find reasonable parameters of the model, such at its size.
Then, these parameters were fixed for the experiment. 
This is to prevent bias towards the loss functions that are novel to this paper. 

In this section, the dense form and tridiagonal form of the operator are considered, but the Jordan form is not considered.
The Jordan form of the operator is considered in the next section.

\subsection{Simple Harmonic Motion}

\begin{figure}
\begin{center}
    \includegraphics[scale=.4]{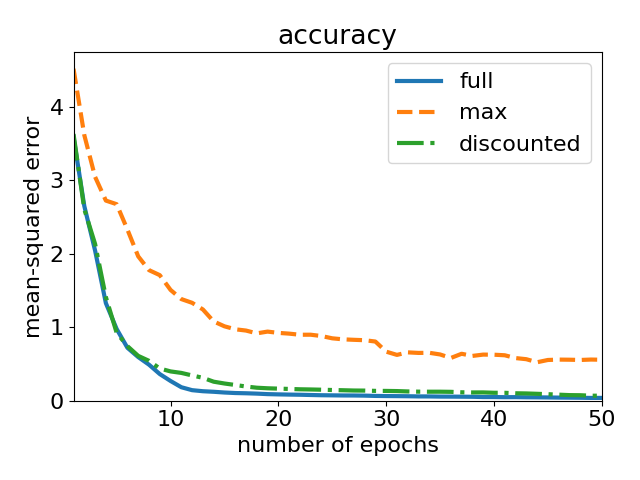}
    \includegraphics[scale=.4]{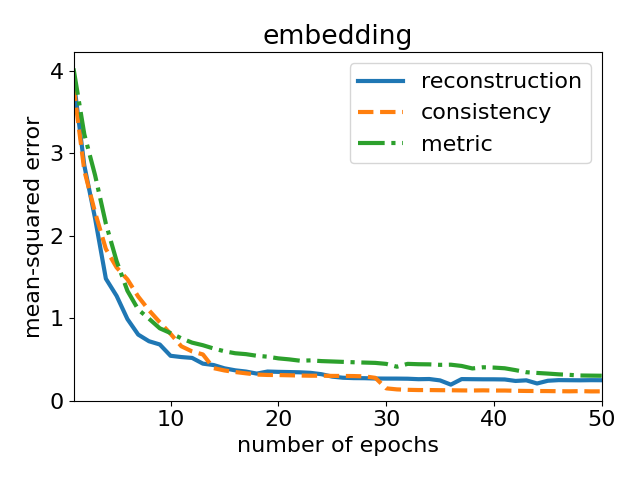}
    \includegraphics[scale=.4]{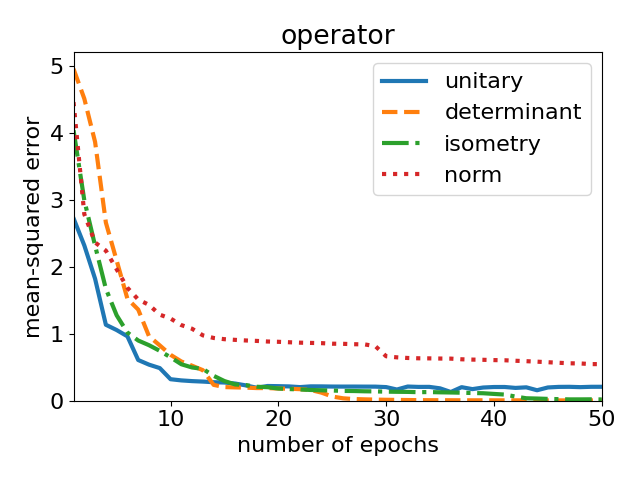}
    \includegraphics[scale=.4]{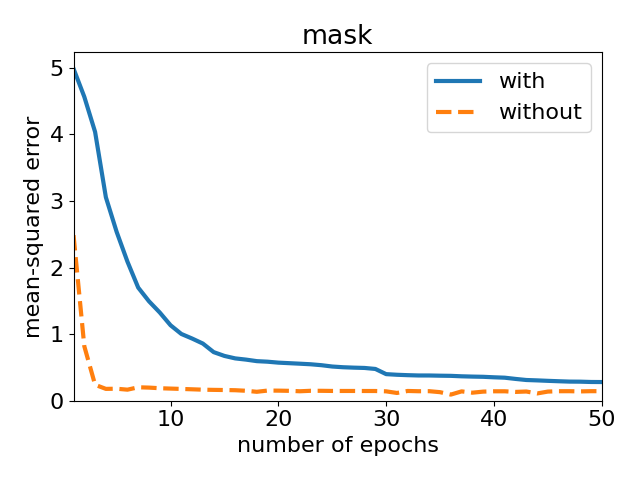}
    \caption{
        The mean effect of the different options of the numerical experiment for simple harmonic motion.
        The top left plot compares the mean effect of the different accuracy loss terms. 
        The top right plot compares the mean effect of the different embedding loss terms. 
        The bottom left plot compares the mean effect of the different operator loss terms. 
        The bottom right plot compares the mean effect of whether or not a mask is used.
        }
    \label{shm_plots}
\end{center}
\end{figure}

\begin{figure}
\begin{center}
    \includegraphics[scale=.5]{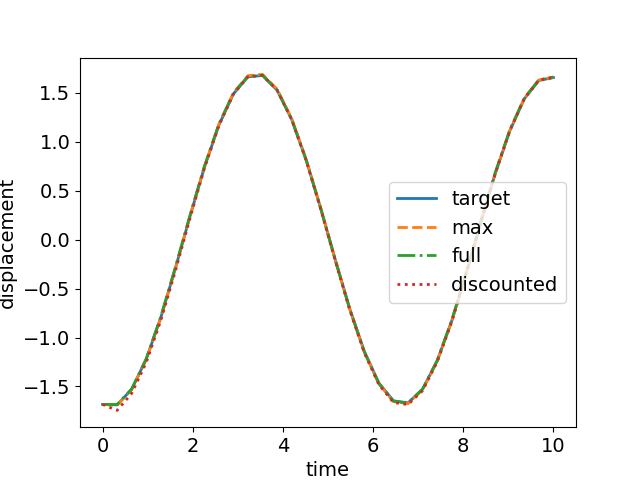}
    \includegraphics[scale=.5]{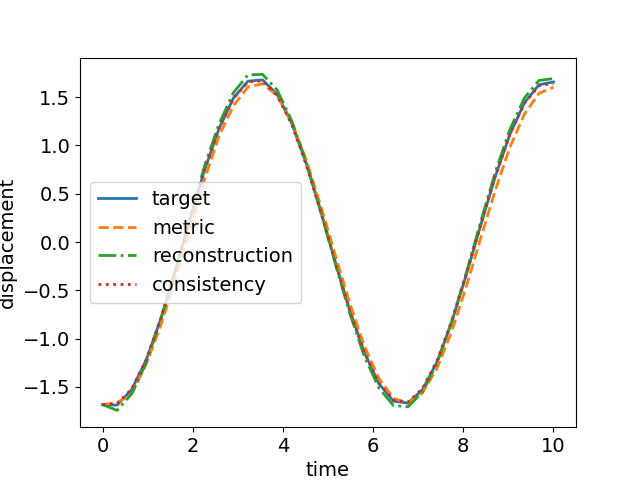}
    \caption{
        Direct comparison of loss terms for the simple harmonic motion. 
        On the left, the accuracy term is varied while the other terms are fixed. 
        On the right, the embedding term is varied while the other terms are fixed. 
        In each comparison, the embedding dimension is $64$, a mask is used, the determinant loss is used, and an auxiliary loss term is not used. 
        In the comparison of the accuracy terms, the reconstruction loss is used. 
        In the comparison of the embedding terms, full accuracy loss is used.
    }
    \label{shm_direct}
\end{center}
\end{figure}

In this numerical experiment, the embedding dimensions $16$, $32$ and $64$ were considered.
Furthermore, all combinations between the accuracy loss terms, the embedding loss terms, the operator loss terms, and whether or not a mask was used so that the operator is tridiagonal were considered, except that the determinant loss is only considered with a mask
Thus, $216$ combinations were considered. 
Figure \ref{shm_plots} shows the mean error for the different options as the number of epochs increases. 
Table \ref{shm_top10} shows the ten best combinations at $20$ epochs.
Table \ref{shm_times} shows the mean relative times for the different options, that is, the mean times for the different options divided by the overall mean time. 
Figure \ref{shm_direct} is direct comparisons between the accuracy terms and also the embedding terms.

\begin{table}
\begin{center}
    \caption{The ten best combinations for the numerical experiment for simple harmonic motion at $20$ epochs. The dimension is the dimension of the latent space.}
    \begin{tabular}{rllllr}
\toprule
error & accuracy & embedding & operator & mask & dimension \\
\midrule
0.000306 & full & reconstruction & unitary & with & 64 \\
0.000318 & full & reconstruction & determinant & with & 16 \\
0.000515 & discounted & reconstruction & unitary & with & 32 \\
0.000535 & max & reconstruction & unitary & with & 64 \\
0.000543 & discounted & reconstruction & determinant & with & 32 \\
0.000575 & full & reconstruction & determinant & with & 32 \\
0.000586 & max & reconstruction & determinant & with & 16 \\
0.000668 & full & reconstruction & determinant & with & 64 \\
0.000729 & discounted & reconstruction & unitary & with & 64 \\
0.000733 & full & reconstruction & unitary & with & 16 \\
\bottomrule
\end{tabular}

    \label{shm_top10}
\end{center}
\end{table}
    
\begin{table}
\begin{center}
    \caption{The mean relative times for the numerical experiment for simple harmonic motion.}
    \begin{tabular}{lr}
\toprule
option & mean time \\
\midrule
full & 1.033703 \\
max & 0.932336 \\
discounted & 1.033961 \\
reconstruction & 1.127289 \\
consistency & 1.023493 \\
metric & 0.849218 \\
unitary & 0.994508 \\
determinant & 1.178594 \\
isometry & 0.899820 \\
norm & 1.016375 \\
with & 1.038539 \\
without & 0.948614 \\
\bottomrule
\end{tabular}

    \label{shm_times}
\end{center}
\end{table}

For the accuracy loss terms, on average, the full accuracy loss and the discounted accuracy loss performed about the same. They both performed significantly better than the maximum accuracy loss.
Five of the best ten combinations used the full accuracy loss, three used the discounted accuracy loss, and two used the maximum accuracy loss.
In the direct comparison, all the accuracy loss terms performed similarly. 

For the embedding loss term, on average, the three loss terms performed about the same. The metric loss performed slightly worse than the reconstruction loss and the consistency loss.
But, all of the best ten combinations used the consistency loss. 
In the direct comparison, the consistency loss does slightly better than the other embedding loss terms.

For the operator loss term, on average, the isometry loss, the unitary loss and the determinant loss performed similarly, and these three terms performed significantly better than the norm loss. 
Five of the best ten combinations used the unitary loss and five used the determinant loss. 

For the mask, on average, having as mask performed much better. 
Furthermore, all of the best ten combinations used a mask.

Some of the same combinations of options with different embedding dimensions appeared in the best ten combinations. 
This demonstrates that these loss combinations are robust. 

For the most part, the mean times for the options are similar. 
The reconstruction loss performed slightly slower than the other embedding loss terms. 
And, the determinant loss term performed slightly slower than the other operator loss terms.

\subsection{Pendulum}

\begin{figure}
\begin{center}
    \includegraphics[scale=.4]{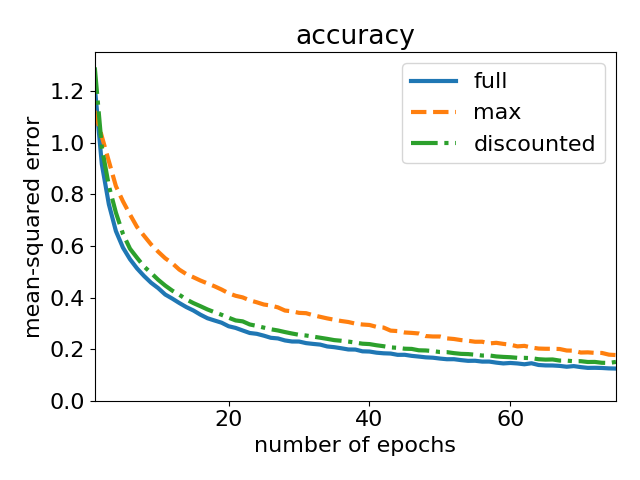}
    \includegraphics[scale=.4]{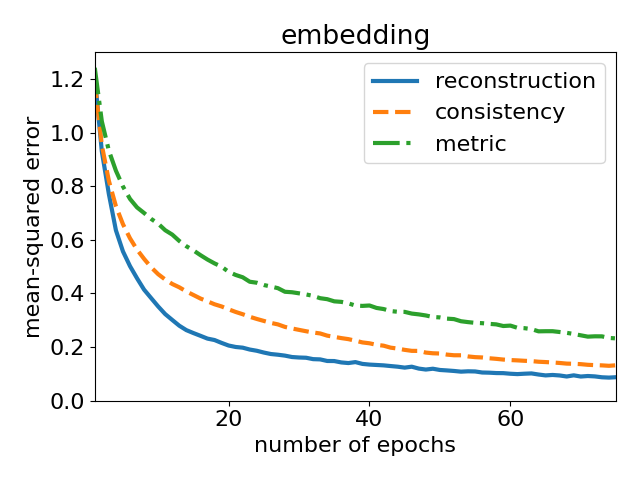}
    \includegraphics[scale=.4]{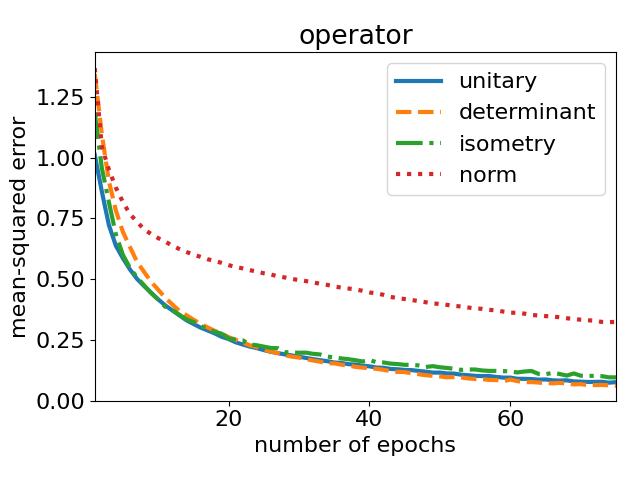}
    \includegraphics[scale=.4]{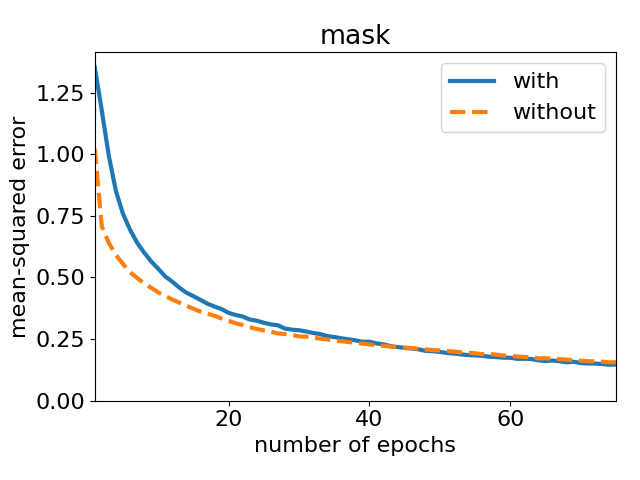}
    \includegraphics[scale=.4]{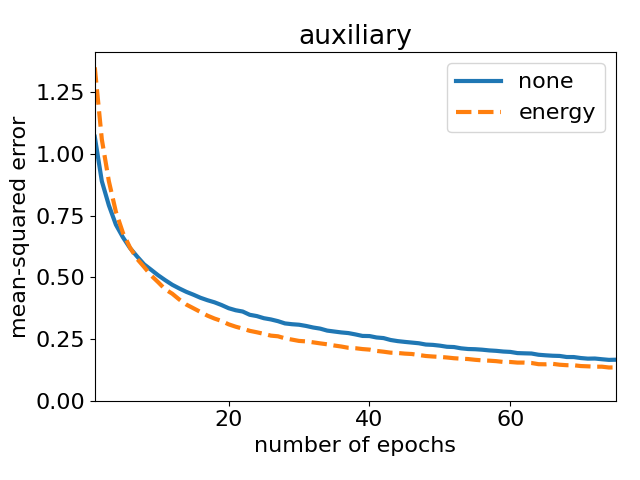}
    \caption{
        The mean effect of the different options of the numerical experiment for the pendulum.
        The top left plot compares the mean effect of the different accuracy loss terms. 
        The top right plot compares the mean effect of the different embedding loss terms. 
        The center left plot compares the mean effect of the different operator loss terms. 
        The center right plot compares the mean effect of whether or not a mask is used.
        The bottom left plot compares the mean effect of the different auxiliary loss terms.
        }
    \label{pendulum_plots}
\end{center}
\end{figure}

In this numerical experiment, the embedding dimensions $32$ and $64$ were considered.
Furthermore, all combinations between the accuracy loss terms, the embedding loss terms, the operator loss terms, whether or not the auxiliary energy-conservation loss term was used and whether or not a mask was used so that the operator is tridiagonal were considered, except that the determinant loss is only considered with a mask
Thus, $288$ combinations were considered. 
Figure \ref{pendulum_plots} shows the mean error for the different options as the number of epochs increases. 
Table \ref{pendulum_top10} shows the ten best combinations at $40$ epochs.
Table \ref{pendulum_times} shows the mean relative times for the different options, that is, the mean times for the different options divided by the overall mean time. 
The energy-conservation loss term is implemented by using 
\[
    \frac{1}{2} \frac{\mathrm{d} \theta}{\mathrm{d} t} + (1-\cos(\theta))
\]
as the energy.

\begin{table}
\begin{center}
    \caption{The ten best combinations for the numerical experiment for the pendulum at $40$ epochs. The dimension is the dimension of the latent space.}
    \begin{tabular}{rlllllr}
\toprule
error & accuracy & embedding & operator & auxiliary & mask & dimension \\
\midrule
0.028013 & discounted & reconstruction & determinant & energy & with & 64 \\
0.028057 & full & reconstruction & unitary & none & with & 32 \\
0.029205 & discounted & reconstruction & isometry & energy & without & 64 \\
0.030154 & full & reconstruction & unitary & energy & with & 32 \\
0.031289 & max & reconstruction & determinant & none & with & 64 \\
0.031841 & max & reconstruction & unitary & none & with & 64 \\
0.032055 & discounted & reconstruction & isometry & none & without & 64 \\
0.032307 & discounted & reconstruction & unitary & none & with & 32 \\
0.032504 & full & reconstruction & unitary & none & with & 64 \\
0.032866 & full & reconstruction & isometry & energy & without & 64 \\
\bottomrule
\end{tabular}

    \label{pendulum_top10}
\end{center}
\end{table}
    
\begin{table}
\begin{center}
    \caption{The mean relative times for the numerical experiment for the pendulum.}
    \begin{tabular}{lr}
\toprule
option & mean time \\
\midrule
full & 1.947080 \\
max & 0.727518 \\
discounted & 0.325402 \\
reconstruction & 1.300984 \\
consistency & 0.946781 \\
metric & 0.752236 \\
unitary & 0.978510 \\
determinant & 1.245490 \\
isometry & 0.974475 \\
norm & 0.924270 \\
none & 1.019804 \\
energy & 0.980196 \\
with & 1.068024 \\
without & 0.909301 \\
\bottomrule
\end{tabular}

    \label{pendulum_times}
\end{center}
\end{table}

For the accuracy loss terms, on average, the full accuracy loss and the discounted accuracy loss performed about the same. They both performed significantly better than the maximum accuracy loss. Four of the best ten combinations used the full accuracy loss, four used the discounted accuracy loss, and two used the maximum accuracy loss.

For the embedding loss term, on average, the reconstruction loss performed the best, followed by the consistency loss and then the metric loss. 
All of the best ten combination used the reconstruction loss.

For the operator loss term, on average, the isometry loss, the unitary loss and the determinant loss performed similarly, and these three terms performed significantly better than the norm loss. 
Five of the best ten combinations used the unitary loss, three used the isometry loss and two used the determinant loss. 

For the mask, on average, having a mask and not having a mask performed similarly.
Seven of the top ten combinations used a mask. 

For the auxiliary loss terms, on average, using the energy-conservation loss term performed slightly better than not using it.
Four of the best ten combinations used the energy-conservation loss term.

Some of the same combinations of options with different embedding dimensions appeared in the best ten combinations. 
This demonstrates that these loss combinations are robust. 

For the most part, the mean times for the options are similar. 
The accuracy loss term performed several factors slower than the other accuracy los terms. 
The reconstruction loss performed slightly slower than the other embedding loss terms. 
And, the determinant loss term performed slightly slower than the other operator loss terms.

\subsection{Lorenz System}

In this numerical experiment, the embedding dimensions 8, 16, 32 and 64 are considered. And, the discount factors 1.0, 0.975 and 0.95 are considered.
All combinations of each of the different types of loss terms of considered and whether or not the mask is used, except that the determinant loss is only considered with a mask. 

\begin{figure}
\begin{center}
    \includegraphics[scale=.4]{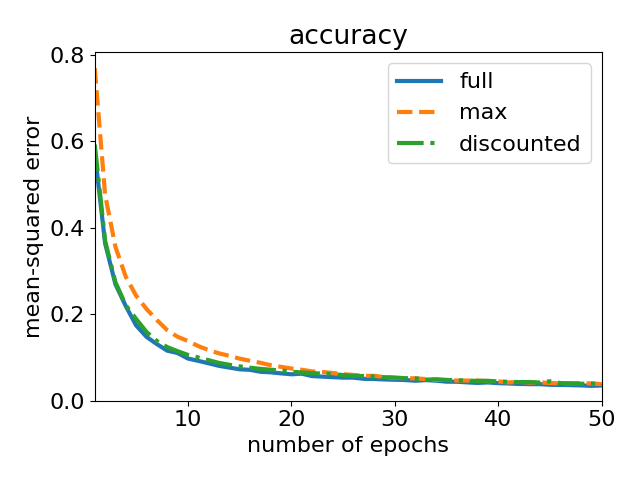}
    \includegraphics[scale=.4]{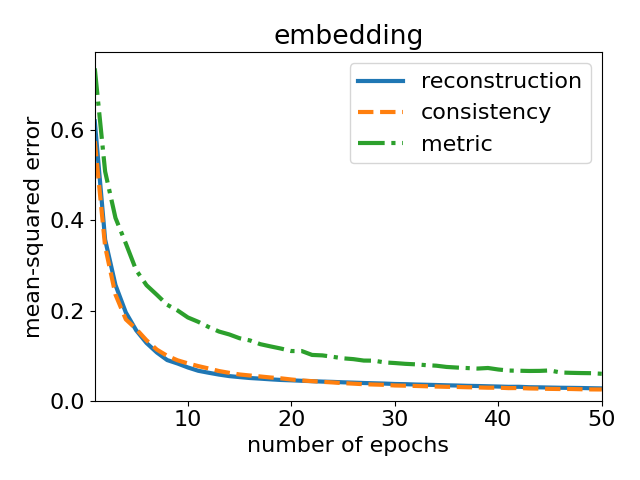}
    \includegraphics[scale=.4]{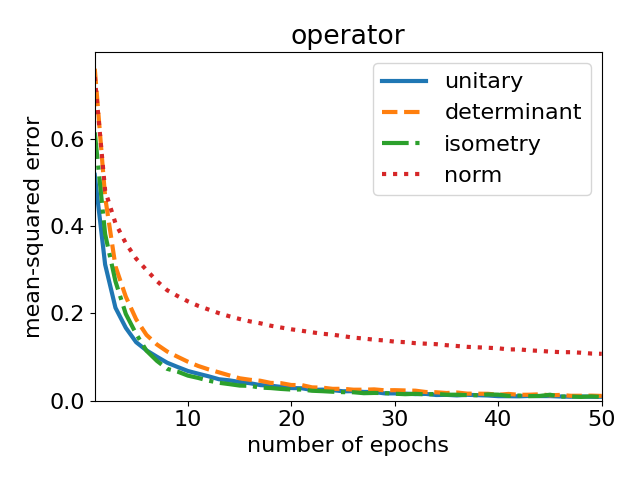}
    \includegraphics[scale=.4]{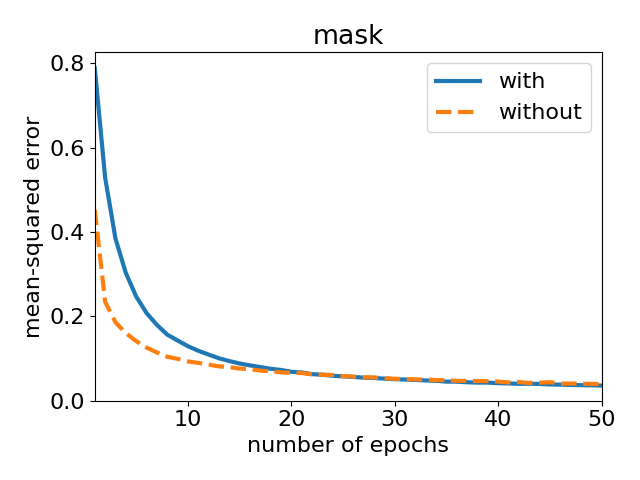}
    \caption{
        The mean effect of the different options of the numerical experiment for the Lorenz system.
        The top left plot compares the mean effect of the different accuracy loss terms. 
        The top right plot compares the mean effect of the different embedding loss terms. 
        The bottom left plot compares the mean effect of the different operator loss terms. 
        The bottom right plot compares the mean effect of whether or not a mask is used.
        }
    \label{lorenz_plots}
\end{center}
\end{figure}

\begin{figure}
\begin{center}
    \includegraphics[scale=.5]{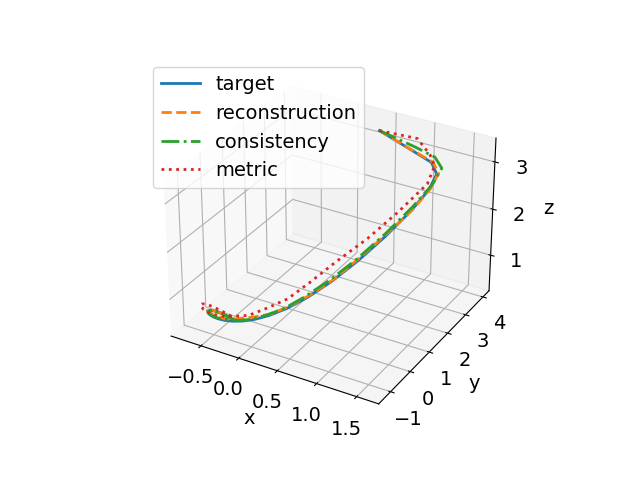}
    \includegraphics[scale=.5]{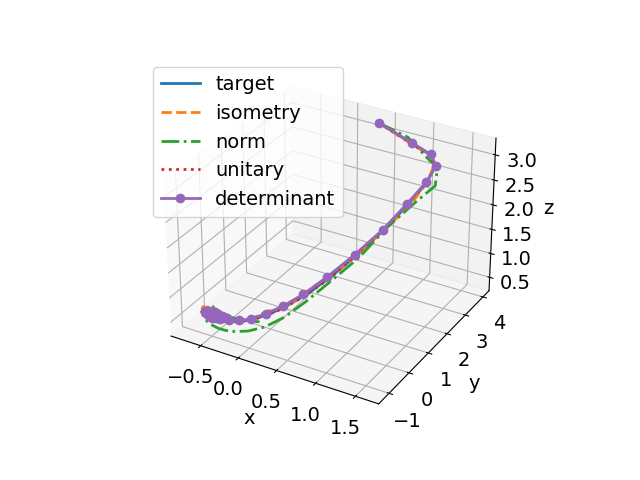}
    \caption{
        Direct comparison of loss terms for the Lorenz system. 
        On the left, the embedding term is varied while the other terms are fixed. 
        On the right, the operator term is varied while the other terms are fixed. 
        In each comparison, the embedding dimension is $64$, a mask is used, the full accuracy loss is used, and an auxiliary loss term is not used. 
        In the comparison of the embedding terms, the unitary loss is used. 
        In the comparison of the operator terms, the reconstruction loss is used. 
    }
    \label{lorenz_direct}
\end{center}
\end{figure}

In this numerical experiment, the embedding dimensions $32$ and $64$ were considered.
Furthermore, all combinations between the accuracy loss terms, the embedding loss terms, the operator loss terms and whether or not a mask was used so that the operator is tridiagonal were considered, except that the determinant loss is only considered with a mask
Thus, $144$ combinations were considered. 
Figure \ref{lorenz_plots} shows the mean error for the different options as the number of epochs increases. 
Table \ref{lorenz_top10} shows the ten best combinations at $30$ epochs.
Table \ref{lorenz_times} shows the mean relative times for the different options, that is, the mean times for the different options divided by the overall mean time. 
Figure \ref{lorenz_direct} is direct comparisons between the embedding terms and also the operator terms.

\begin{table}
\begin{center}
    \caption{The ten best combinations for the numerical experiment for the Lorenz system at $30$ epochs. The dimension is the dimension of the latent space.}
    \begin{tabular}{rllllr}
\toprule
error & accuracy & embedding & operator & mask & dimension \\
\midrule
0.001249 & full & reconstruction & unitary & with & 64 \\
0.001505 & discounted & reconstruction & determinant & with & 64 \\
0.001605 & discounted & reconstruction & unitary & with & 64 \\
0.001691 & full & reconstruction & determinant & with & 64 \\
0.002287 & max & reconstruction & unitary & with & 64 \\
0.002292 & max & reconstruction & determinant & with & 64 \\
0.002741 & full & consistency & unitary & with & 64 \\
0.002878 & full & reconstruction & isometry & without & 64 \\
0.002879 & full & consistency & isometry & with & 64 \\
0.002900 & max & consistency & isometry & with & 64 \\
\bottomrule
\end{tabular}

    \label{lorenz_top10}
\end{center}
\end{table}
    
\begin{table}
\begin{center}
    \caption{The mean relative times for the numerical experiment for the Lorenz system.}
    \begin{tabular}{lr}
\toprule
option & mean time \\
\midrule
full & 1.859016 \\
max & 0.726902 \\
discounted & 0.414082 \\
reconstruction & 1.273235 \\
consistency & 0.895752 \\
metric & 0.831013 \\
unitary & 1.039408 \\
determinant & 1.156576 \\
isometry & 0.983032 \\
norm & 0.899272 \\
with & 1.053328 \\
without & 0.928897 \\
\bottomrule
\end{tabular}

    \label{lorenz_times}
\end{center}
\end{table}

For the accuracy loss terms, on average, all the accuracy terms performed similarly.
Five of the best ten combinations used the full accuracy loss, three used the maximum accuracy loss, and two used the discounted accuracy loss.

For the embedding loss term, on average, the metric loss performed significantly worse than the reconstruction loss and the consistency loss. 
Seven of the best ten combinations used the reconstruction loss, and three used the consistency loss. 
In the direct comparison, the metric loss does worse than the other embedding loss terms.

For the operator loss term, on average, the isometry loss, the unitary loss and the determinant loss performed similarly, and these three terms performed significantly better than the norm loss. 
Four of the best ten combinations used the unitary loss, three used the isometry loss and three used the determinant loss. 
In the direct comparison, the norm loss does worse than the other operator loss terms.

For the mask, on average, having a mask and not having a mask performed similarly.
Furthermore, nine of the top ten combinations used a mask. 

In all of the top ten combinations, the embedding dimension was $64$.
This demonstrates that the model size was more important than the options. 
It is likely that the other embedding dimension were too small for the Lorenz system. 

For the most part, the mean times for the options are similar. 
The accuracy loss term performed several factors slower than the other accuracy los terms. 
The reconstruction loss performed slightly slower than the other embedding loss terms. 
And, the determinant loss term performed slightly slower than the other operator loss terms.

\subsection{Heat Equation}

\begin{figure}
\begin{center}
    \includegraphics[scale=.4]{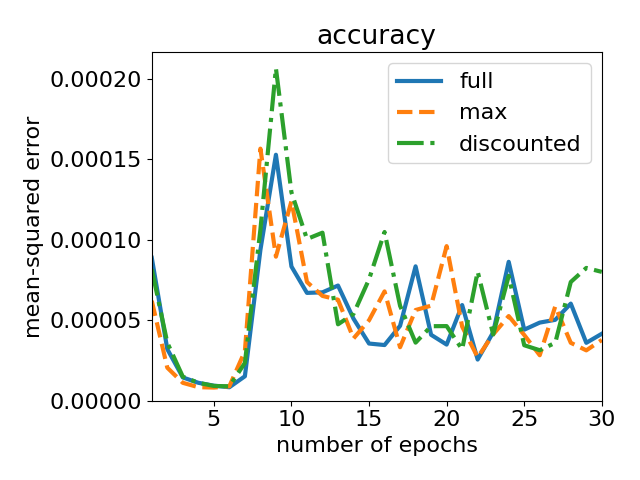}
    \includegraphics[scale=.4]{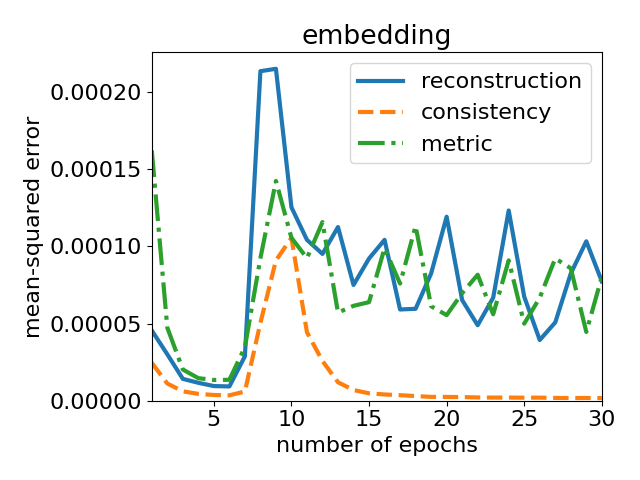}
    \includegraphics[scale=.4]{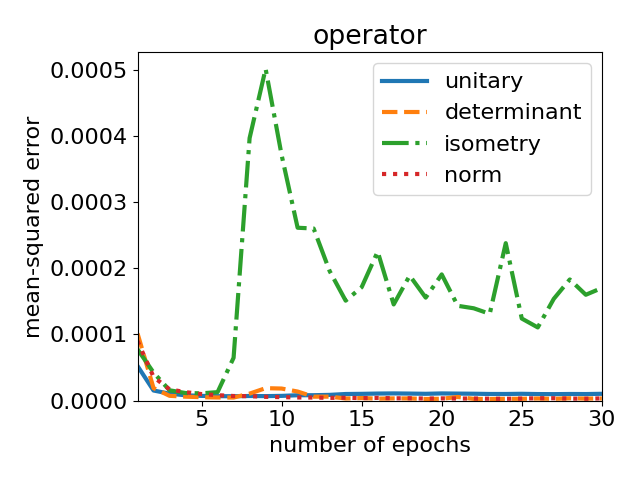}
    \includegraphics[scale=.4]{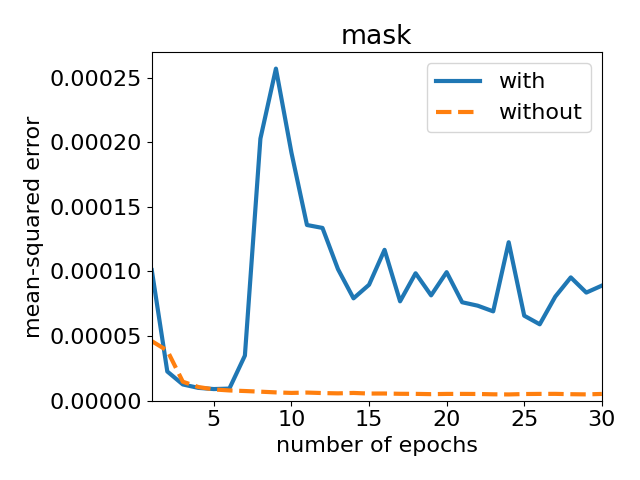}
    \includegraphics[scale=.4]{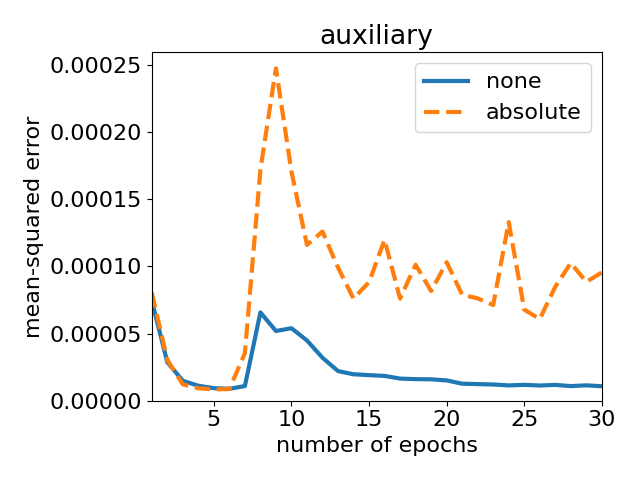}
    \caption{
        The mean effect of the different options of the numerical experiment for the heat equation.
        The top left plot compares the mean effect of the different accuracy loss terms. 
        The top right plot compares the mean effect of the different embedding loss terms. 
        The center left plot compares the mean effect of the different operator loss terms. 
        The center right plot compares the mean effect of whether or not a mask is used.
        The bottom left plot compares the mean effect of the different auxiliary loss terms.
        }
    \label{heat_plots}
\end{center}
\end{figure}

In this numerical experiment, the embedding dimensions $64$ and $128$ were considered.
Furthermore, all combinations between the accuracy loss terms, the embedding loss terms, the operator loss terms, whether or not the auxiliary absolute-max loss term was used and whether or not a mask was used so that the operator is tridiagonal were considered, except that the determinant loss is only considered with a mask
Thus, $288$ combinations were considered. 
Figure \ref{heat_plots} shows the mean error for the different options as the number of epochs increases. 
Table \ref{heat_top10} shows the ten best combinations at 20 epochs.
Table \ref{heat_times} shows the mean relative times for the different options, that is, the mean times for the different options divided by the overall mean time. 

\begin{table}
\begin{center}
    \caption{The ten best combinations for the numerical experiment for the heat equation at $20$. The dimension is the dimension of the latent space.}
    \begin{tabular}{rlllllr}
\toprule
error & accuracy & embedding & operator & auxiliary & mask & dimension \\
\midrule
1e-7 & discounted & consistency & determinant & none & with & 128 \\
1e-6 & full & consistency & determinant & none & with & 128 \\
1e-6 & discounted & consistency & isometry & none & without & 128 \\
1e-6 & full & consistency & determinant & none & with & 64 \\
1e-6 & discounted & consistency & determinant & none & with & 64 \\
1e-6 & max & consistency & norm & none & with & 64 \\
1e-6 & full & consistency & norm & absolute & with & 128 \\
1e-6 & max & consistency & determinant & absolute & with & 128 \\
1e-6 & full & consistency & isometry & none & without & 128 \\
1e-7 & full & consistency & isometry & none & without & 64 \\
\bottomrule
\end{tabular}

    \label{heat_top10}
\end{center}
\end{table}

\begin{table}
\begin{center}
    \caption{The mean relative times for the numerical experiment for the heat equation.}
    \begin{tabular}{lr}
\toprule
option & mean time \\
\midrule
full & 0.979201 \\
max & 1.003339 \\
discounted & 1.017461 \\
reconstruction & 1.100878 \\
consistency & 0.989917 \\
metric & 0.909205 \\
unitary & 0.957837 \\
determinant & 1.193389 \\
isometry & 0.962771 \\
norm & 0.982697 \\
none & 0.994160 \\
absolute & 1.005840 \\
with & 1.035062 \\
without & 0.953251 \\
\bottomrule
\end{tabular}

    \label{heat_times}
\end{center}
\end{table}

For the accuracy loss terms, on average, all the accuracy terms performed similarly.
Five of the best ten combinations used the full accuracy loss, three used the discounted accuracy loss, and two used the maximum accuracy loss.

For the embedding loss term, on average, the consistency loss performed significantly better than the metric loss and the reconstruction loss.
Furthermore, all of the best ten combination used the consistency loss.

For the operator loss term, on average, the isometry loss performed significantly worse than the other operator loss terms.
Five of the best ten combinations used the determinant loss, three used the isometry loss and two used the norm loss. 

For the mask, on average, not having as mask performed much better. 
But, seven of the best ten combinations used a mask.

Some of the same combinations of options with different embedding dimensions appeared in the best ten combinations. 
This demonstrates that these loss combinations are robust. 

For the most part, the mean times for the options are similar. 
The determinant loss term performed slightly slower than the other operator loss terms.

\subsection{Wave Equation}

\begin{figure}
\begin{center}
    \includegraphics[scale=.4]{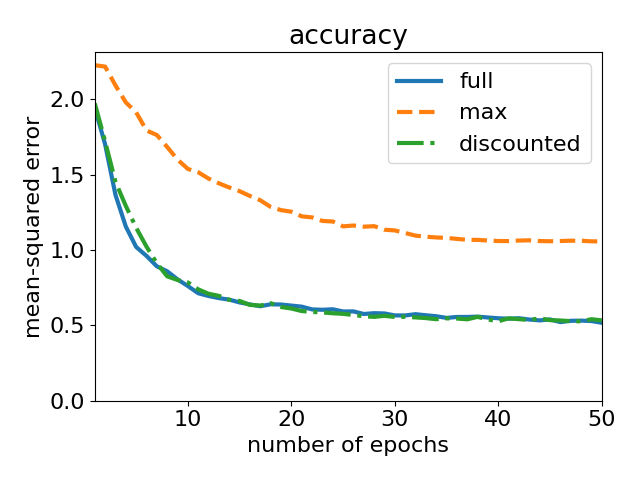}
    \includegraphics[scale=.4]{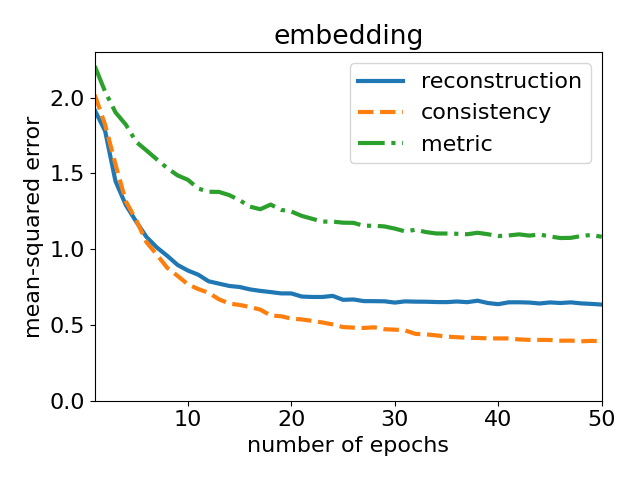}
    \includegraphics[scale=.4]{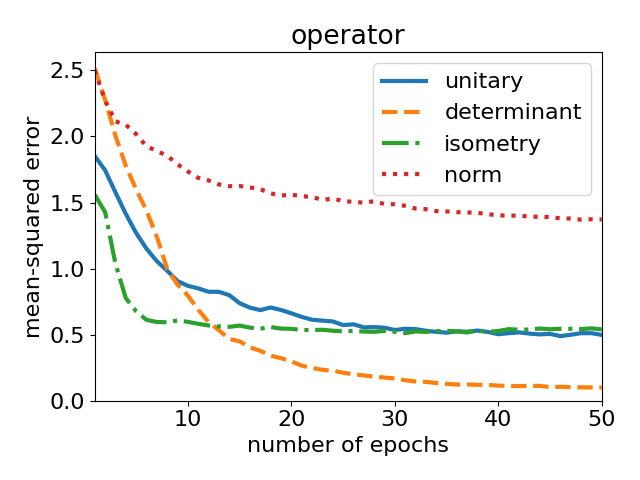}
    \includegraphics[scale=.4]{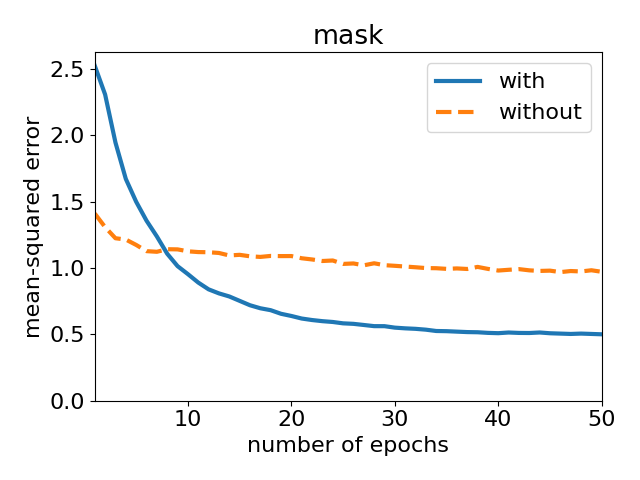}
    \includegraphics[scale=.4]{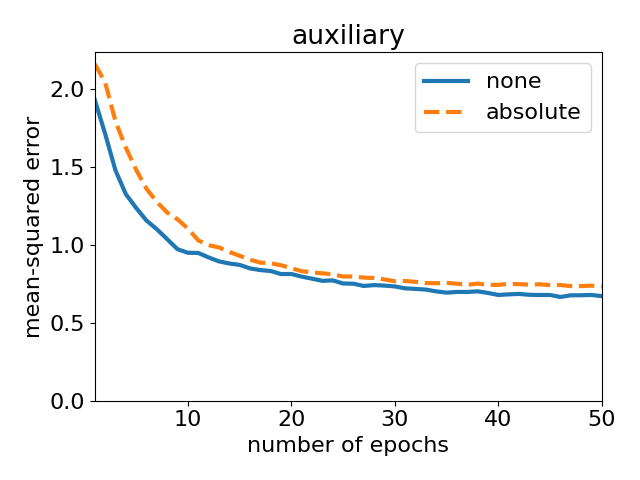}
    \caption{
        The mean effect of the different options of the numerical experiment for the wave equation.
        The top left plot compares the mean effect of the different accuracy loss terms. 
        The top right plot compares the mean effect of the different embedding loss terms. 
        The center left plot compares the mean effect of the different operator loss terms. 
        The center right plot compares the mean effect of whether or not a mask is used.
        The bottom left plot compares the mean effect of the different auxiliary loss terms.
        }
    \label{wave_plots}
\end{center}
\end{figure}

\begin{figure}
\begin{center}
    \includegraphics[scale=.5]{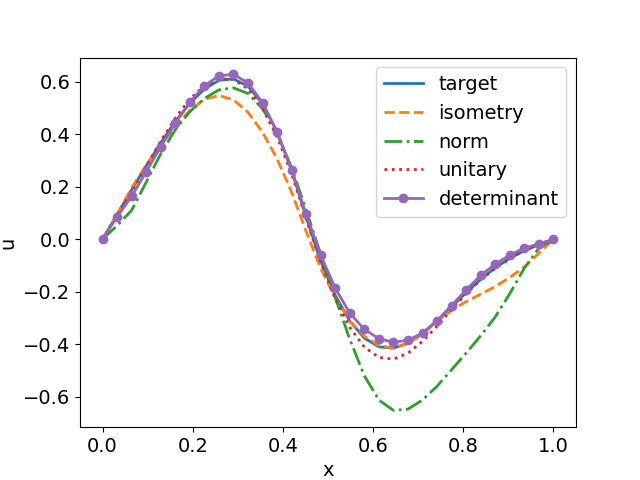}
    \includegraphics[scale=.5]{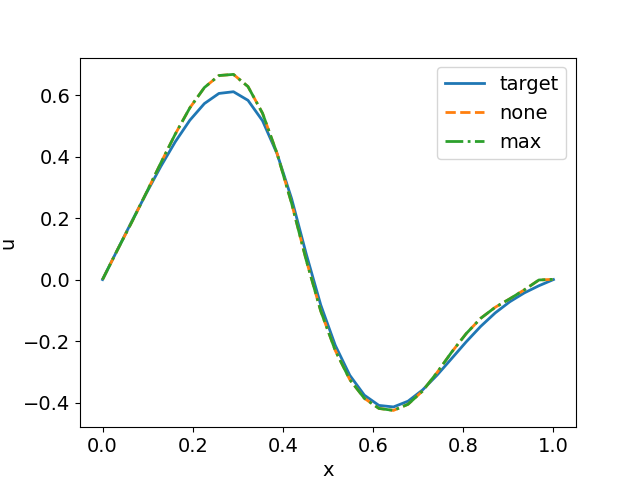}
    \caption{
        Direct comparison of loss terms for the wave equation. 
        Each plot is a time slice of the solution at a time later than the start time. 
        On the left, the operator term is varied while the other terms are fixed. 
        On the right, the auxiliary term is varied while the other terms are fixed. 
        In each comparison, the embedding dimension is $64$, a mask is used, the full accuracy loss is used, and the reconstruction loss is used.
        In the comparison of the operator terms, an auxiliary loss term is not used
        In the comparison of the auxiliary loss term, the unitary loss is used. 
    }
    \label{wave_direct}
\end{center}
\end{figure}

In this numerical experiment, the embedding dimensions $256$ was considered.
Furthermore, all combinations between the accuracy loss terms, the embedding loss terms, the operator loss terms, whether or not the absolute max loss was used and whether or not a mask was used so that the operator is tridiagonal were considered, except that the determinant loss is only considered with a mask
Thus, $288$ combinations were considered. 
Figure \ref{wave_plots} shows the mean error for the different options as the number of epochs increases. 
Table \ref{wave_top10} shows the ten best combinations at 30 epochs.
Table \ref{wave_times} shows the mean relative times for the different options, that is, the mean times for the different options divided by the overall mean time. 
Figure \ref{wave_direct} is direct comparisons between the embedding terms and also the operator terms.

\begin{table}
\begin{center}
    \caption{The ten best combinations for the numerical experiment with the wave equation at $30$ epochs. The dimension is the dimension of the latent space.}
    \begin{tabular}{rlllllr}
\toprule
error & accuracy & embedding & operator & auxiliary & mask & dimension \\
\midrule
0.004519 & discounted & reconstruction & unitary & none & with & 256 \\
0.004697 & full & consistency & isometry & none & with & 256 \\
0.005081 & full & reconstruction & unitary & none & with & 256 \\
0.005167 & discounted & consistency & determinant & none & with & 128 \\
0.005582 & full & consistency & unitary & none & with & 256 \\
0.005680 & full & reconstruction & unitary & none & with & 128 \\
0.005806 & discounted & consistency & unitary & none & with & 256 \\
0.005870 & discounted & reconstruction & unitary & none & with & 128 \\
0.005918 & discounted & consistency & unitary & none & with & 128 \\
0.005954 & full & reconstruction & determinant & none & with & 256 \\
\bottomrule
\end{tabular}

    \label{wave_top10}
\end{center}
\end{table}
    
\begin{table}
\begin{center}
    \caption{The mean relative times for the numerical experiment for the wave equation.}
    \begin{tabular}{lr}
\toprule
option & mean time \\
\midrule
full & 1.018608 \\
max & 1.005451 \\
discounted & 0.975941 \\
reconstruction & 1.088658 \\
consistency & 0.985603 \\
metric & 0.925739 \\
unitary & 0.922116 \\
determinant & 1.321372 \\
isometry & 0.894227 \\
norm & 1.022971 \\
none & 0.997791 \\
absolute & 1.002209 \\
with & 1.084602 \\
without & 0.887197 \\
\bottomrule
\end{tabular}

    \label{wave_times}
\end{center}
\end{table}

For the accuracy loss terms, on average, the full accuracy loss and the discounted accuracy loss performed about the same. They both performed significantly better than the maximum accuracy loss. 
Five of the best ten combinations used the full accuracy loss, and five used the discounted accuracy loss.

For the embedding loss term, on average, the consistency loss performed the best, followed by the reconstruction loss and then the metric loss. 
Five of the best ten combinations used the consistency loss, and five used the reconstruction loss. 

For the operator loss term, on average, the determinant loss performed the best, followed by the isometry loss and unitary loss, and then the norm loss. 
Seven of the best ten combinations used the unitary loss, two used the determinant loss, and one used the isometry loss. 
In the direct comparison, the norm loss performed worse than the other operator loss terms, and the determinant loss performed the best.

For the auxiliary loss terms, on average, using or not using the absolute maximum loss performed similarly. 
But, all of the best ten combinations did not use an auxiliary loss term. 
In the direct comparison, using or not using the absolute maximum loss performed similarly.

For the mask, on average, having a mask performed better.
Furthermore, all of the top ten combinations used a mask. 

Some of the same combinations of options with different embedding dimensions appeared in the best ten combinations. 
This demonstrates that these loss combinations are robust. 

For the most part, the mean times for the options are similar. 
The determinant loss term performed slightly slower than the other operator loss terms.

\subsection{Burger's Equation}

\begin{figure}
\begin{center}
    \includegraphics[scale=.4]{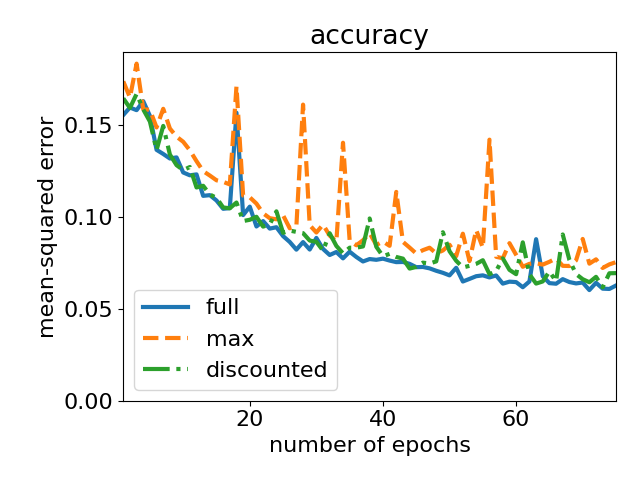}
    \includegraphics[scale=.4]{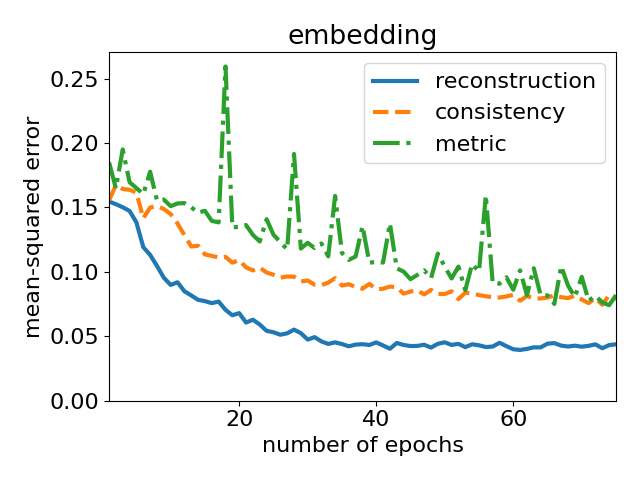}
    \includegraphics[scale=.4]{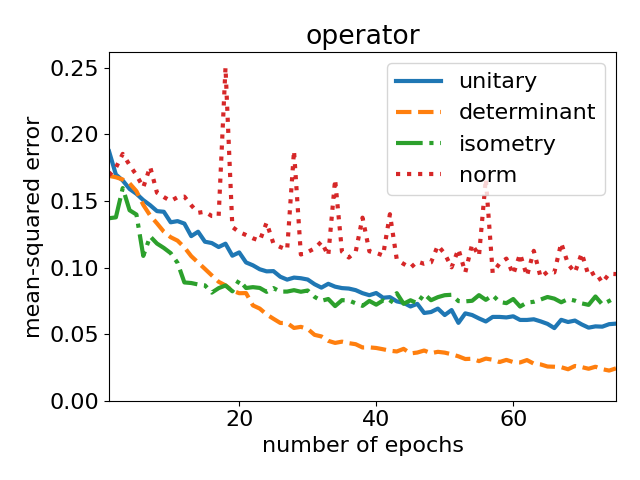}
    \includegraphics[scale=.4]{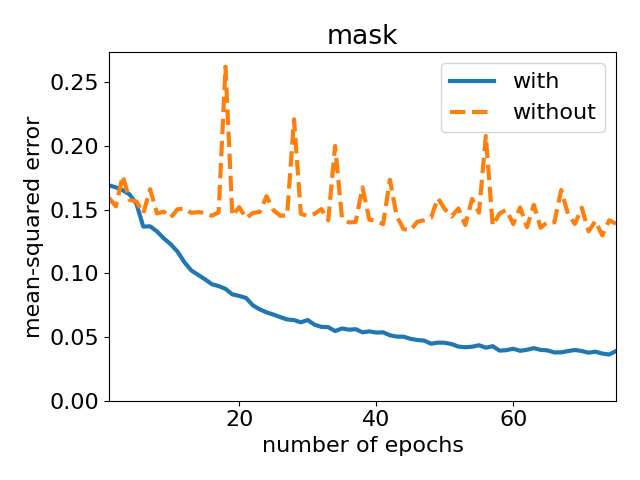}
    \includegraphics[scale=.4]{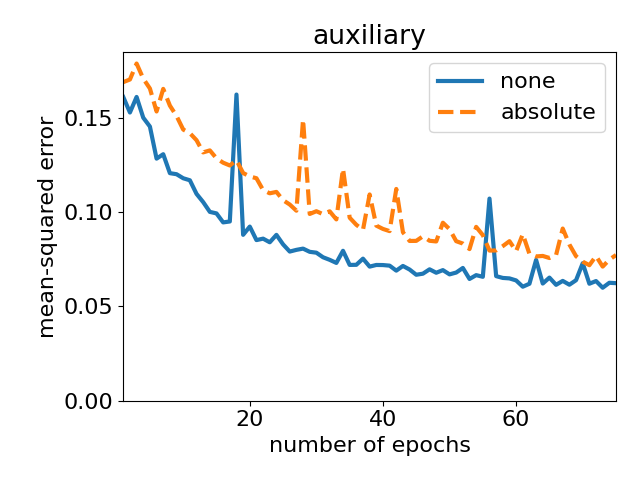}
    \caption{
        The mean effect of the different options of the numerical experiment for Burger's equation.
        The top left plot compares the mean effect of the different accuracy loss terms. 
        The top right plot compares the mean effect of the different embedding loss terms. 
        The center left plot compares the mean effect of the different operator loss terms. 
        The center right plot compares the mean effect of whether or not a mask is used.
        The bottom left plot compares the mean effect of the different auxiliary loss terms.
        }
    \label{burgers_plots}
\end{center}
\end{figure}

\begin{figure}
\begin{center}
    \includegraphics[scale=.5]{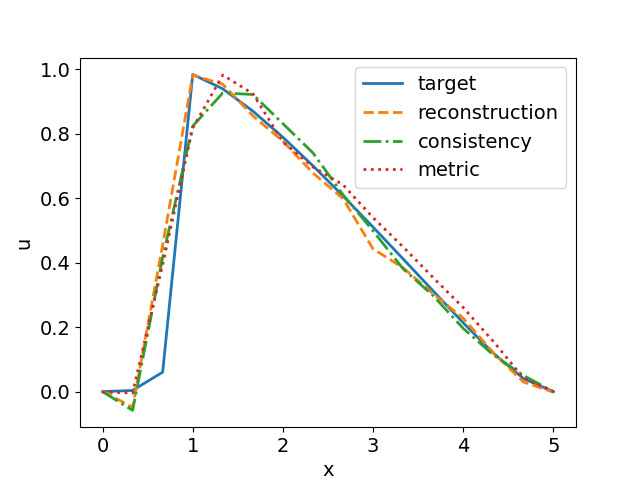}
    \includegraphics[scale=.5]{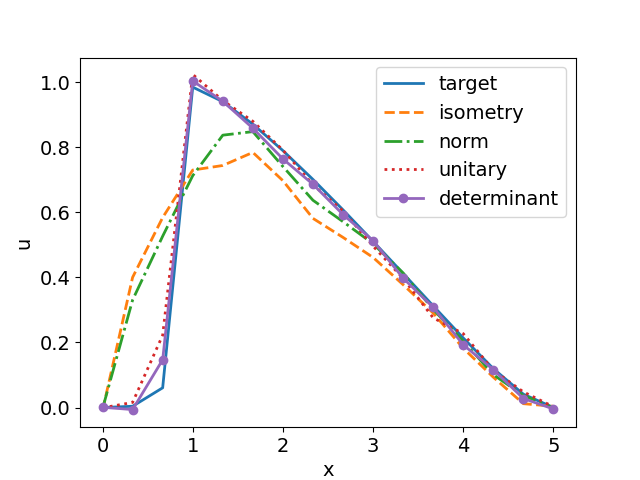}
    \caption{
        Direct comparison of loss terms for Burger's equation. 
        Each plot is a time slice of the solution at a time later than the start time. 
        On the left, the embedding term is varied while the other terms are fixed. 
        On the right, the operator term is varied while the other terms are fixed. 
        In each comparison, the embedding dimension is $64$, a mask is used, the full accuracy loss is used, and an auxiliary loss term is not used. 
        In the comparison of the embedding terms, the unitary loss is used. 
        In the comparison of the operator terms, the reconstruction loss is used. 
    }
    \label{burgers_direct}
\end{center}
\end{figure}

In this numerical experiment, the embedding dimensions $512$ and $1024$ were considered.
Furthermore, all combinations between the accuracy loss terms, the embedding loss terms, the operator loss terms, whether or not the auxiliary absolute-max loss term was used and whether or not a mask was used so that the operator is tridiagonal were considered, except that the determinant loss is only considered with a mask
Thus, $288$ combinations were considered. 
Figure \ref{burgers_plots} shows the mean error for the different options as the number of epochs increases. 
Table \ref{burgers_top10} shows the ten best combinations at $50$ epochs.
Table \ref{burgers_times} shows the mean relative times for the different options, that is, the mean times for the different options divided by the overall mean time. 
Figure \ref{burgers_direct} is direct comparisons between the embedding terms and also the operator terms.

\begin{table}
\begin{center}
    \caption{The ten best combinations for the numerical experiment for Burger's equation motion at $50$ epochs. The dimension is the dimension of the latent space.}
    \begin{tabular}{rlllllr}
\toprule
error & accuracy & embedding & operator & auxiliary & mask & dimension \\
\midrule
0.002115 & full & consistency & unitary & none & with & 512 \\
0.002416 & full & reconstruction & unitary & none & with & 512 \\
0.002751 & full & consistency & isometry & none & with & 512 \\
0.002813 & full & reconstruction & determinant & none & with & 512 \\
0.003033 & full & consistency & determinant & none & with & 512 \\
0.003074 & discounted & consistency & unitary & none & with & 512 \\
0.003309 & discounted & consistency & isometry & none & with & 512 \\
0.003746 & discounted & consistency & determinant & none & with & 512 \\
0.004211 & full & consistency & unitary & none & with & 1024 \\
0.004310 & max & consistency & isometry & none & with & 512 \\
\bottomrule
\end{tabular}

    \label{burgers_top10}
\end{center}
\end{table}
        
\begin{table}
\begin{center}
    \caption{The mean relative times for the numerical experiment for Burger's equations.}
    \begin{tabular}{lr}
\toprule
option & mean time \\
\midrule
full & 0.996650 \\
max & 0.997402 \\
discounted & 1.005948 \\
reconstruction & 1.030242 \\
consistency & 0.987900 \\
metric & 0.981858 \\
unitary & 2.283492 \\
determinant & 0.987301 \\
isometry & 0.361504 \\
norm & 0.361354 \\
none & 0.999075 \\
absolute & 1.000925 \\
with & 0.999741 \\
without & 1.000345 \\
\bottomrule
\end{tabular}

    \label{burgers_times}
\end{center}
\end{table}

For the accuracy loss terms, on average, the full accuracy loss and the discounted accuracy loss performed about the same. They both performed slightly better than the maximum accuracy loss. 
Seven of the best ten combinations used the full accuracy loss, and three used the discounted accuracy loss.

For the embedding loss term, on average, the reconstruction loss performed the best, followed by the consistency loss and then the metric loss. 
Eight of the best ten combinations used the consistency loss, and two used the reconstruction loss. 
In the direct comparison, the reconstruction loss performed the best. 

For the operator loss term, on average, the determinant loss performed the best, followed by the isometry loss and unitary loss, and then the norm loss. 
Four of the best ten combinations used the unitary loss, three used the determinant loss, and three used the isometry loss. 
In the direct comparison, the unitary loss and determinant loss performed better than the isometry loss and the norm loss. 

For the auxiliary loss terms, on average, not using the absolute maximum loss performed better. 
Furthermore, all of the best ten combinations did not use an auxiliary loss term. 

For the mask, on average, having a mask performed better.
Furthermore, all of the top ten combinations used a mask. 

Only one combination of options was repeated in the best ten combinations. 
And, it was the only best ten combination that did not use the $512$ for the embedding dimension. 
The embedding dimension $1024$ may have been too big for such an early epoch. 

For the most part, the mean times for the options are similar. 
The unitary loss performed slightly slower than the other operator loss terms by over a factor of $2$.

\subsection{Conclusion}

The full accuracy loss is the most robust accuracy loss term, and it is recommend. 
If the model is obviously struggling with initial learning, then it may be beneficial to try the discounted accuracy loss.

The reconstruction loss and the consistency loss performed similarly overall. 
One did better for some equations, and the other did better for other equations. 
It may be beneficial to try both of these terms in practice. 
But, either term is probably safe to use.

The tridiagonal operator was more robust than the dense operator. 
The unitary loss and the determinant loss are the most robust operator loss terms. 
The operator form and loss term is further studied in the next section. 

Overall the auxiliary loss terms did not increase performance significantly. 
It is not normally recommended to use them. 
But, for some systems, it may be beneficial to experiment with a physics-informed loss term that is specific to the system. 

In each experiment, the speeds of each option were similar. 
The differences in speed do not seem significant enough to be a factor is choosing the loss functions. 

The biggest differences were that the reconstruction loss was slightly slower than the other encoder loss terms and that the determinant loss was slightly slower than the other operator loss terms. 
Intuitively, it makes sense that the reconstruction loss is slower than the other encoder loss terms because it requires the most additional computations. 
In these experiments, the reconstruction loss was trained using the data at each time step.
But, in the prediction of many time steps, it might be reasonable to train it on less data to reduce the computational cost.

The determinant loss was slower than the other operator loss terms. 
Intuitively, it is slower because it cannot be parallelized.
But, it scales linearly, and this should not become a problem with any model size. 

In these experiments, the operator was a dense matrix in the backend computations. 
But, a custom layer could be created for the tridiagonal form. 
Then, multiplication can be defined on this layer so that it scales linearly with the dimension.

\section{Loss for and Form of the Operator}\label{sec-operator}
The experiments in this section compares the loss terms for and operator forms for Koopman autoencoders.
Unlike the experiments in the previous section, these experiments are for many epochs. 
For the experiments in this section, the full accuracy loss and the reconstruction loss are used because these were demonstrated to be the most robust in the previous section. 
Furthermore, other than possibly an operator loss term, another loss term is not used. 
Furthermore, because the previous section already studied the times for various loss terms, this is not studied in this section. 

For each equation, all combinations between the operator forms and operator loss terms are considered, expect that the determinant loss is not considered with the dense operator form. 
Furthermore, unlike in the previous section, the possibility of not using an operator loss term is included.

\subsection{Numerical Experiments}

Table \ref{pendulum_op} is the results for the equation for the pendulum. 
The most accurate combination was the tridiagonal form with the unitary loss.
The tridiagonal form of the operator was demonstrated to be the most robust form. 

\begin{table}[H]
\begin{center}
    \caption{Results for the pendulum}
    \begin{tabular}{lll}
    error & operator form & operator loss \\
    \toprule
    1.66353e-3 & dense & none \\
    6.6002e-4 & dense & isometry \\
    6.31303e-2 & dense & norm \\
    1.89346e-2 & dense & unitary \\
    3.8495e-4 & tridiagonal & none \\
    7.4554e-4 & tridiagonal & isometry \\
    4.59331e-4 & tridiagonal & norm \\
    1.2317e-4 & tridiagonal & unitary \\
    3.8344e-4 & tridiagonal & determinant \\
    1.55339e-1 & jordan & none \\
    2.63119e-4 & jordan & isometry \\
    1.96144e-2 & jordan & norm \\
    2.5279e-2 & jordan & unitary \\
    2.42197e-1 & jordan & determinant \\
    \end{tabular}
    \label{pendulum_op}
\end{center}
\end{table}

\begin{figure}[H]
\begin{center}
    \includegraphics[scale=.5]{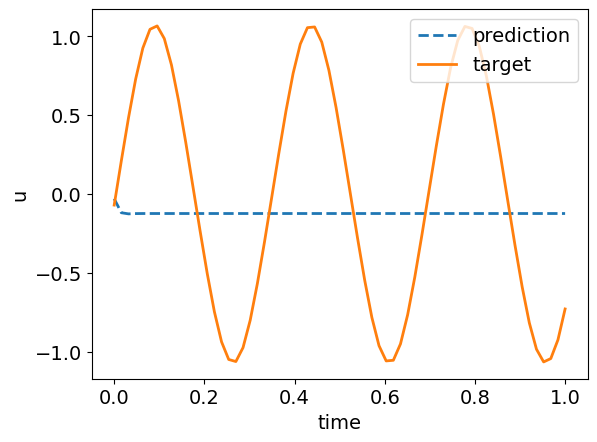}
    \includegraphics[scale=.5]{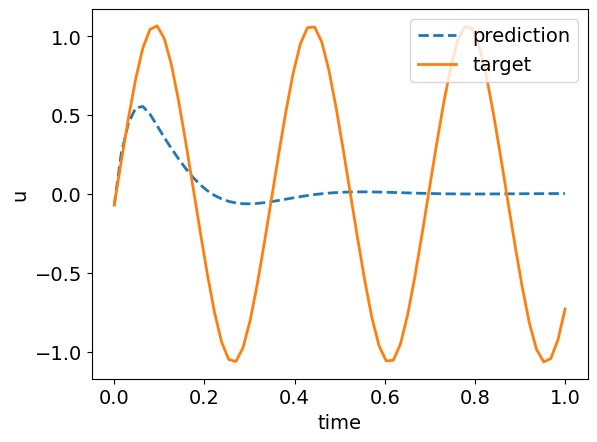}
    \includegraphics[scale=.5]{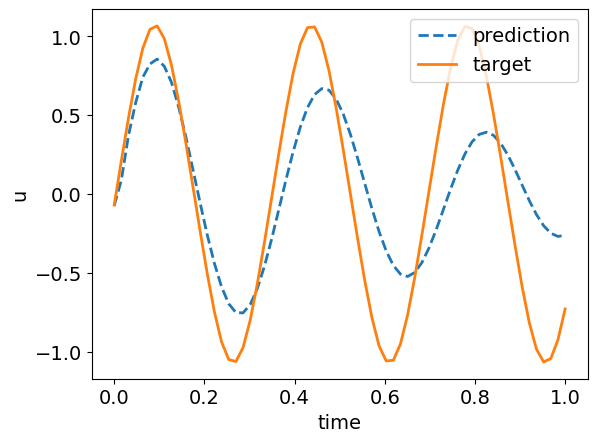}
    \includegraphics[scale=.5]{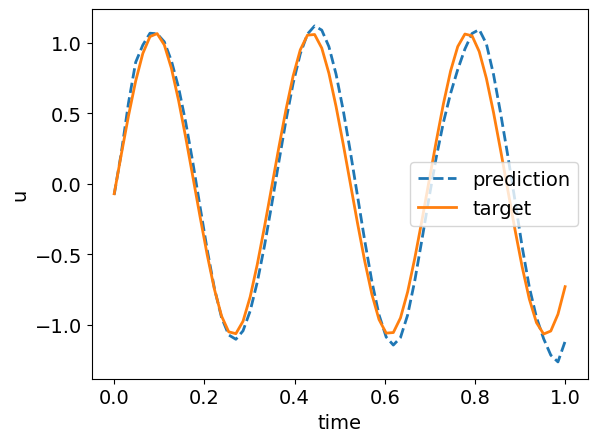}
    \caption{
        Progress learning the equation for the pendulum: The model uses the tridiagonal form of the operator and the determinant loss. The frames are at $1$, $10$, $50$, and $200$ epochs. The scales are normalized. 
    }
\end{center}
\end{figure}

Table \ref{lorenz_op} is the results for the equation for the pendulum. 
The most accurate combination was the Jordan form without an operator loss term. 
Unlike the other experiments, the lack of an operator loss term does well. 

\begin{table}[H]
\begin{center}
    \caption{Results for the Lorenz system}
    \begin{tabular}{lll}
    error & operator form & operator loss \\
    \toprule
    7.312e-3 & dense & none \\
    1.390e-2 & dense & isometry \\
    8.495e-2 & dense & norm \\
    1.342e-2 & dense & unitary \\
    6.368e-3 & tridiagonal & none \\
    1.586e-2 & tridiagonal & isometry \\
    1.182e-1 & tridiagonal & norm \\
    7.386e-3 & tridiagonal & unitary \\
    6.679e-3 & tridiagonal & determinant \\
    2.114e-3 & jordan & none \\
    5.266e-3 & jordan & isometry \\
    3.989e-2 & jordan & norm \\
    8.151e-3 & jordan & unitary \\
    1.961e-2 & jordan & determinant \\
    \end{tabular}
    \label{lorenz_op}
\end{center}
\end{table}

Table \ref{fluid_op} is the results for the equation for the fluid attractor equation. 
The most accurate combination was the tridiagonal form with the determinant loss.
The tridiagonal form of the operator was demonstrated to be the most robust form.

\begin{table}[H]
\begin{center}
    \caption{Results for the fluid attractor equation}
    \begin{tabular}{lll}
    error & operator form & operator loss \\
    \toprule
    2.47e-6 & dense & none \\
    7.02e-5 & dense & isometry \\
    1.30e-2 & dense & norm \\
    1.040e-5 & dense & unitary \\
    3.79-6 & tridiagonal & none \\
    2.52e-5 & tridiagonal & isometry \\
    1.80e-2 & tridiagonal & norm \\
    3.69e-6 & tridiagonal & unitary \\
    1.99e-6 & tridiagonal & determinant \\
    2.020e-2 & jordan & none \\
    2.08e-5 & jordan & isometry \\
    1.67e-2 & jordan & norm \\
    7.05-6 & jordan & unitary \\
    2.20-2 & jordan & determinant \\
    \end{tabular}
    \label{fluid_op}
\end{center}
\end{table}

\begin{figure}[H]
\begin{center}
    \includegraphics[scale=.5]{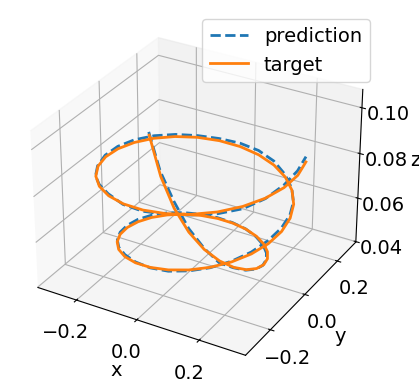}
    \hspace{1cm}
    \includegraphics[scale=.5]{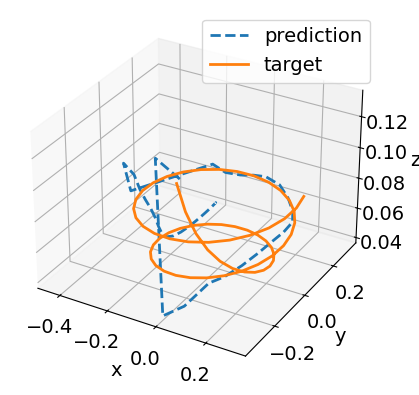}
    \caption{
        Comparison of two operator loss terms for the fluid attractor equation: Both plots are of the trajectories of the target and the prediction. On the left, the model uses the tridiagonal form of the operator and the determinant loss. On the right, the model uses the dense form of the operator and the norm loss. These predictions are at $2000$ epochs.  
    }
\end{center}
\end{figure}

Table \ref{burgers_op} is the results for the equation for the fluid attractor equation. 
The most accurate combination was the tridiagonal form without an unitary loss. 
The tridiagonal form of the operator was demonstrated to be the most robust form.

\begin{table}[H]
\begin{center}
    \caption{Results for Burger's equation}
    \begin{tabular}{lll}
    error & operator form & operator loss \\
    \toprule
    3.269e-3 & dense & none \\
    4.753e-3 & dense & isometry \\
    3.048e-2 & dense & norm \\
    2.824e-3 & dense & unitary \\
    1.234e-3 & tridiagonal & none \\
    3.666e-3 & tridiagonal & isometry \\
    1.998e-2 & tridiagonal & norm \\
    1.375e-3 & tridiagonal & unitary \\
    1.703e-3 & tridiagonal & determinant \\
    2.470e-2 & jordan & none \\
    2.082e-3 & jordan & isometry \\
    2.20952e-2 & jordan & norm \\
    2.268e-3 & jordan & unitary \\
    2.1471e-2 & jordan & determinant \\
    \end{tabular}
    \label{burgers_op}
\end{center}
\end{table}

Table \ref{kdv_op} is the results for the equation for the fluid attractor equation. 
The most accurate combination was the tridiagonal form with the unitary loss.
The tridiagonal form of the operator was demonstrated to be the most robust form.

\begin{table}[H]
\begin{center}
    \caption{Results for the KdV equation}
    \begin{tabular}{lll}
    error & operator form & operator loss \\
    \toprule
    2.531e-1 & dense & none \\
    3.164e-1 & dense & isometry \\
    1.699e-1 & dense & norm \\
    2.838e-2 & dense & unitary \\
    1.579e-1 & tridiagonal & none \\
    2.410e-1 & tridiagonal & isometry \\
    7.294e-1 & tridiagonal & norm \\
    4.467e-2 & tridiagonal & unitary \\
    4.807e-2 & tridiagonal & determinant \\
    1.421e-1 & jordan & none \\
    1.254 & jordan & isometry \\
    1.012e-1 & jordan & norm \\
    9.378e-1 & jordan & unitary \\
    5.0974e-1 & jordan & determinant \\
    \end{tabular}
    \label{kdv_op}
\end{center}
\end{table}

\subsection{Conclusion} 

The best combination varied from equation to equation. Overall, the tridiagonal form of the operator is the most robust. Furthermore, the unitary loss term was the most robust operator loss term. It is recommended to use the tridiagonal form with the unitary loss term. 

\section{Conclusion}\label{sec-conclusion}
The full accuracy loss term is the most robust accuracy loss term. 
The reconstruction loss term and the consistency loss terms are the most robust encoding loss terms. 
The tridiagonal form of the operator is the most robust operator form. 
And, the unitary loss term is the most robust operator loss term.
It is recommended to use the full accuracy loss term, the reconstruction loss term, and the unitary loss term with the tridiagonal form of the operator.

\section*{Code Availability}
The code for the numerical experiments is publicly available at \url{https://gitlab.com/dustin_lee/neural-operators}.
All experiments were done in Python\index{Python}. 
Neural networks were implemented with PyTorch\index{PyTorch} \cite{paszke2017automatic, pointer2019programming, lippe2024uvadlc} and Lightning\index{Lightning} \cite{lightning}. 
Hyperparameter configuration was done using Hydra\index{Hydra} \cite{Yadan2019Hydra}.

\section*{Declaration of Competing Interest}
The authors declare that they have no known competing financial interests or personal relationships that could have appeared to influence the work reported in this paper.

\section*{Acknowledgments}
Guang Lin would like to thank the support of the National Science Foundation (DMS-2053746, DMS-2134209, ECCS-2328241, CBET-2347401 and OAC-2311848), and U.S.~Department of Energy (DOE) Office of Science Advanced Scientific Computing Research program DE-SC0023161, and DOE–Fusion Energy Science, under grant number: DE-SC0024583.

\printbibliography

\end{document}